# Specifying Nonspecific Evidence

Johan Schubert

*Division of Information System Technology, Department of Command and Control Warfare Technology, National Defence Research Establishment, S-172 90 Stockholm, Sweden*

In an earlier article [J. Schubert, "On nonspecific evidence," *Int. J. Intell. Syst.* **8**(6), 711−725 (1993)] we established within Dempster-Shafer theory a criterion function called the metaconflict function. With this criterion we can partition into subsets a set of several pieces of evidence with propositions that are weakly specified in the sense that it may be uncertain to which event a proposition is referring. Each subset in the partitioning is representing a separate event. The metaconflict function was derived as the plausibility that the partitioning is correct when viewing the conflict in Dempster's rule within each subset as a newly constructed piece of metalevel evidence with a proposition giving support against the entire partitioning. In this article we extend the results of the previous article. We will not only find the most plausible subset for each piece of evidence as was done in the earlier article. In addition we will specify each piece of nonspecific evidence, in the sense that we find to which events the proposition might be referring, by finding the plausibility for every subset that this piece of evidence belong to the subset. In doing this we will automatically receive indication that some evidence might be false. We will then develop a new methodology to exploit these newly specified pieces of evidence in a subsequent reasoning process. This will include methods to discount evidence based on their degree of falsity and on their degree of credibility due to a partial specification of affiliation, as well as a refined method to infer the event of each subset. © 1996 John Wiley & Sons, Inc.

## I. INTRODUCTION

When we are reasoning under uncertainty in an environment of several possible events we may find some pieces of evidence that are not only uncertain but may also have propositions that are weakly specified in the sense that it may not be certain to which event a proposition is referring. In addition our own domain knowledge regarding the current number of events may be uncertain. In this situation we must make sure that we do not by mistake combine the pieces of evidence that are referring to different events.

The methodology to handle and specify nonspecific pieces of evidence was developed as a part of a multiple-target tracking algorithm for an antisubmarine intelligence analysis system.[1,2] In this application a sparse flow of intelligence reports arrives at the analysis system. These reports may originate from several different unconnected sensor systems. The reports carry a proposition about the occurrence of a submarine at a specified time and place, a probability of the truthfulness of the report and may contain additional information such as velocity, direction and type of submarine.

When there are several submarines we want to separate the intelligence reports into subsets according to which submarine they are referring to. We will then analyze the reports for each submarine separately. However, the intelligence reports are never labeled as to which submarine they are referring to. Thus, it is not possible to directly differentiate between two different submarines using two intelligence reports.

Instead we will use the conflict between the propositions of two intelligence reports as a probability that the two reports are referring to different submarines. This probability is the basis for separating intelligence reports into subsets.

The cause of the conflict can be nonfiring sensors placed between the positions of the two reports, the required velocity to travel between the positions of the two reports at their respective times in relation to the assumed velocity of the submarines, etc.

The general idea is this. If we receive several pieces of evidence about several different and separate events and the pieces of evidence are mixed up, we want to sort all the pieces of evidence according to which event they are referring to. Thus, we partition the set of all pieces of evidence into subsets where each subset refers to a particular event. In Figure 1 these subsets are denoted by $\chi_i$. Here, 13 pieces of evidence are partitioned into four subsets. When the number of subsets is uncertain there will also be a "domain conflict" which is a conflict between the current number of subsets and domain knowledge. The partition is then simply an allocation of all pieces of evidence to the different events. Since these events do not have anything to do with each other, we will analyze them separately.

Now, if it is uncertain to which event some pieces of evidence is referring we have a problem. It could then be impossible to know directly if two different pieces of evidence are referring to the same event. We do not know if we should put them into the same subset or not. This problem is then a problem of organization. Evidence from different problems that we want to analyze are unfortunately mixed up and we are having some problem separating it.

To solve this problem, we can use the conflict in Dempster's rule when all pieces of evidence within a subset are combined, as an indication of whether these pieces of evidence belong together. The higher this conflict is, the less credible that they belong together.

Let us create an additional piece of evidence for each subset where the proposition of this additional piece of evidence states that this is not an "adequate partition." Let the proposition take a value equal to the conflict of the combination within the subset. These new pieces of evidence, one regarding each subset, reason about the partition of the original evidence. Just so we do not confuse them with the original evidence, let us call all this evidence *metalevel evidence* and let us say that its combination and the analysis of that combination take place on the *metalevel* (Fig. 1).

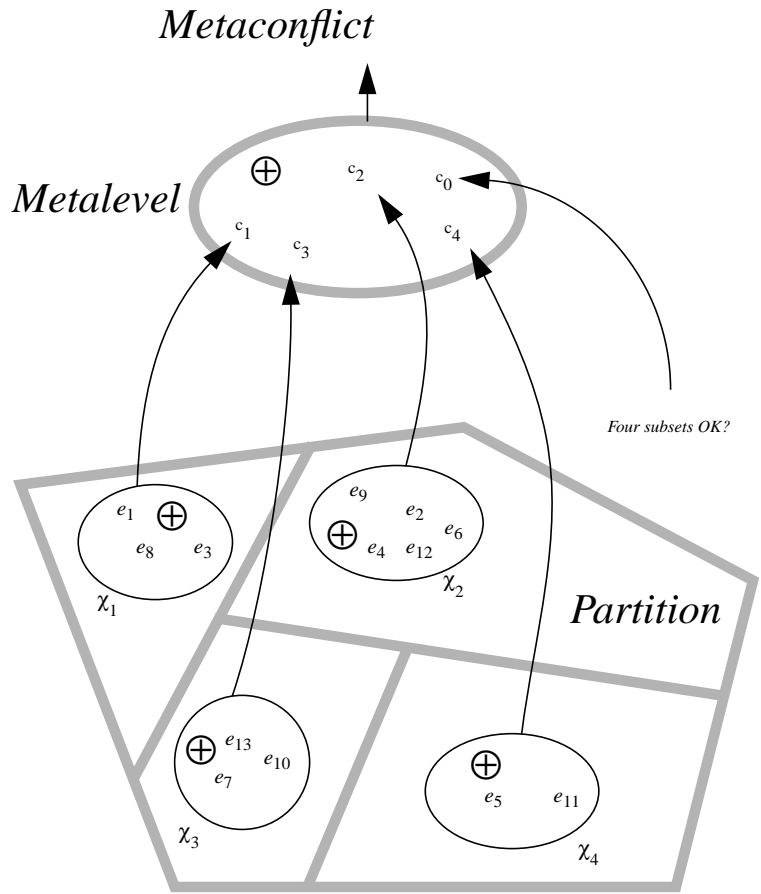

**Figure 1.** The Conflict in each subset of the partition becomes a piece of evidence at the metalevel.

In the combination of all pieces of metalevel evidence, one regarding each subset, we only receive support stating that this is not an "adequate partition." We may call this support a *metaconflict*. The smaller this support is, the more credible the partition. Thus, the most credible partition is the one that minimizes the metaconflict.

We know that it is not possible to separate the pieces of evidence based only on their proposition. Instead we prefer to separate it based on their conflicts. Since the conflict measures the lack of compatibility between several pieces of evidence, and pieces of evidence referring to different events tend to be more incompatible than pieces of evidence referring to the same event, it is an obvious choice as a distance measure in some cluster algorithm. The idea of using the conflict in Dempster's rule as distance measure between pieces of evidence was first suggested by Lowrance and Garvey.[3]

In an earlier article[4] we established, within the framework of Dempster-Shafer theory,[5-9] the criterion function of overall conflict called the metaconflict function. With this criterion we can partition evidence with weakly specified propositions into subsets, each subset representing a separate event. These events should be handled independently.

To make a separation of evidence possible, every proposition's action part must be supplemented with an event part describing to which event the proposition is referring. If the proposition is written as a conjunction of literals or disjunctions, then one literal or disjunction concerns which event the proposition is referring to. This is the event part. The remainder of the proposition is called the action part. An example from our earlier article illustrates the terminology:

> Let us consider the burglaries of two bakers' shops at One and Two Baker Street, event 1 ($E_1$) and event 2 ($E_2$), i.e., the number of events is known to be two. One witness hands over a piece of evidence, specific with respect to event 1, with the proposition: "The burglar at One Baker Street," event part: $E_1$, "was probably brown haired ($B$)," action part: $B$. A second anonymous witness hands over a nonspecific piece of evidence with the proposition: "The burglar at Baker Street," event part: $E_1$, $E_2$, "might have been red haired ($R$)," action part: $R$. That is, for example:

| evidence 1: | evidence 2: |
|---|---|
| proposition: | proposition: |
| action part: $B$ | action part: $R$ |
| event part: $E_1$ | event part: $E_1,E_2$ |
| $m(B) = 0.8$ | $m(R) = 0.4$ |
| $m(\Theta) = 0.2$ | $m(\Theta) = 0.6$ |

We will have a conflict between two pieces of evidence in the same subset in two different situations. First, we have a conflict if the proposition action parts are conflicting regardless of the proposition event parts since they are presumed to be referring to the same event. Secondly, if the proposition event parts are conflicting then, regardless of the proposition action parts, we have a conflict with the presumption that they are referring to the same event.

The metaconflict used to partition the set of evidence is derived[4] as the plausibility that the partitioning is correct when the conflict in each subset is viewed as a piece of metalevel evidence against the partitioning of the set of evidence, $\chi$, into the subsets, $\chi_i$. We have a simple frame of discernment on the metalevel $\Theta = \{\text{AdP}, \neg\text{AdP}\}$, where AdP is short for "adequate partition," and a basic probability assignment (bpa) from each subset $\chi_i$ assigning support to a proposition against the partitioning:

$$m_{\chi_i}(\neg\text{AdP}) \triangleq \text{Conf}(\{e_j | e_j \in \chi_i\}),$$

$$m_{\chi_i}(\Theta) \triangleq 1 - \text{Conf}(\{e_j | e_j \in \chi_i\})$$

where $e_j$ is the $j$th piece of evidence and $\{e_j | e_j \in \chi_i\}$ is the evidence belonging to subset $\chi_i$ and Conf($\cdot$) is the conflict, $k$, in Dempster's rule. Also, we have a bpa concerning the domain resulting from a probability distribution about the number of subsets, $E$, conflicting with the actual current number of subsets, $\#\chi$. This bpa also assigns support to a proposition against the partitioning:

$$m_D(\neg \text{AdP}) \triangleq \text{Conf}(\{E, \#\chi\}),$$

$$m_D(\Theta) \triangleq 1 - \text{Conf}(\{E, \#\chi\}).$$

The combination of these by Dempster's rule give us the following plausibility of the partitioning:

$$\text{Pls(AdP)} = (1 - m_D(\neg \text{AdP})) \cdot \prod_{i=1}^{r} (1 - m_{\chi_i}(\neg \text{AdP})).$$

The difference, one minus the plausibility of a partitioning, will be called the metaconflict of the partitioning. The metaconflict function can then be defined as:

DEFINITION.[4] *Let the* metaconflict function,

$$Mcf(r, e_1, e_2, ..., e_n) \triangleq 1 - (1 - c_0) \cdot \prod_{i=1}^{r} (1 - c_i),$$

*be the conflict against a partitioning of n pieces of evidence of the set $\chi$ into r disjoint subsets $\chi_i$ where*

$$c_0 = \sum_{i \neq r} m(E_i)$$

*is the conflict between the hypothesis that there are r subsets and our prior belief about the number of subsets with $m(E_i)$ being the prior support given to the fact there are i subsets and*

$$c_i = \sum_{\substack{I \\ \cap I = \varnothing}} \prod_{e_j^k \in I} m_j(e_j^k)$$

*is the conflict in subset i, where $\cap I$ is the intersection of all elements in I, $I = \{e_j^k | e_j \in \chi_i\}$ is a set of one focal element from the support function of each piece of evidence $e_j$ in $\chi_i$ and $e_j^k$ is the kth focal element of $e_j$.*

Thus, $|I| = |\chi_i|$ and

$$|\{I\}| = \prod_{e_j \in \chi_i} |e_j|$$

where $|e_j|$ is the number of focal elements of $e_j$.

We are here only considering the case where the function $m(\cdot)$ in the calculation of $c_0$ is a probability function.

Two theorems are derived[4] to be used in the separation of evidence into subsets by an iterative minimization of the metaconflict function. By using these theorems we are able to reason about the optimal estimate of number of events, when the actual number of events may be uncertain, as well as the optimal partition of nonspecific evidence for any fixed number of events. These two theorems will also be useful in a process for specifying pieces of evidence by observing changes in the metaconflict when moving a single piece of evidence between different subsets.

THEOREM 1. *For all j with $j < r$, if $m(E_j) < m(E_r)$ then min $Mcf(r, e_1, e_2, ..., e_n) < $ min $Mcf(j, e_1, e_2, ..., e_n)$.*

This theorem states that an optimal partitioning for $r$ subsets is always better than the other solutions with fewer than $r$ subsets if the basic probability assignment for $r$ subsets is greater than the basic probability assignment for the fewer subsets.

THEOREM 2. *For all j, if min $Mcf(r, e_1, e_2, ..., e_n) < \sum_{i \neq j} m(E_i)$ then min $Mcf(r, e_1, e_2, ..., e_n) < $ min $Mcf(j, e_1, e_2, ..., e_n)$.*

Theorem 2 states that an optimal partitioning for some number of subsets is always better than the other solutions for any other number of subsets when the domain part of the metaconflict function is greater than the total metaconflict of the present partitioning.

In Section II we derive a bpa supporting that a piece of evidence is not belonging to a certain subset. This is done by observing the cluster conflict variations when a piece of evidence is moved out from a subset, or when, after that, it is brought into another subset, or by observing the domain conflict variation when it is put into a newly created subset by itself. With this derived bpa we find the support for each piece of evidence and every subset. In Section III we specify all pieces of evidence by combining the bpa's from different subsets regarding a particular piece of evidence and then calculate the plausibility for each subset that this particular piece of evidence belongs to the subset. The entire derivations of Sections II and III are found in Appendices I and II. In the combination of all bpa's in Section III we receive support for a false statement that a piece of evidence does not belong to any of the subsets and cannot be placed in a new subset by itself. We discuss how this situation can be handled in Section IV. In Section V we study the usefulness of the now specified evidence. Obviously, a piece of evidence that can belong to several different subsets is not so useful and should not be allowed to strongly influence a subsequent reasoning process within a subset. We then describe an improved method of finding the event represented by a subset (Sec. VI). Finally, we illustrate the methodology by an example of some "bakers' shops burglaries" and make a comparison of the refined methodology advocated in this article with the simpler approach in our earlier article[4] (Sec. VII).

## II.   EVIDENCE ABOUT EVIDENCE

### A.   Evidence From Cluster Conflict Variations

A conflict in a subset can be interpreted as a piece of metalevel evidence that there is at least one piece of evidence that does not belong to the subset. Thus, we can refine the basic probability assignment from subset $\chi_i$ assigning support to a proposition against the partitioning,

$$m_{\chi_i}(\neg \text{AdP}) = c_i,$$
$$m_{\chi_i}(\Theta) = 1 - c_i$$

to

$$m_{\chi_i}(\exists j . e_j \notin \chi_i) = c_i,$$
$$m_{\chi_i}(\Theta) = 1 - c_i.$$

Let us observe one piece of evidence $e_q$ in $\chi_i$. If $e_q$ is taken out from $\chi_i$ the conflict $c_i$ in $\chi_i$ decreases to $c_i^*$. This decrease in conflict $c_i - c_i^*$ can be interpreted as follows: there exists some metalevel evidence indicating that $e_q$ does not belong to $\chi_i$,

$$m_{\Delta\chi_i}(e_q \notin \chi_i),$$
$$m_{\Delta\chi_i}(\Theta),$$

and the remainder of the conflict $c_i^*$ is metalevel evidence that there is at least one other piece of evidence $e_j$, $j \neq q$, that does not belong to $\chi_i - \{e_q\}$,

$$m_{\chi_i - \{e_q\}}(\exists j \neq q . e_j \notin (\chi_i - \{e_q\})) = c_i^*,$$
$$m_{\chi_i - \{e_q\}}(\Theta) = 1 - c_i^*.$$

We will derive the basic probability number of $e_q \notin \chi_i$, $m_{\Delta\chi_i}(e_q \notin \chi_i)$, by stating that the belief in the proposition that there is at least one piece of evidence that does not belong to $\chi_i$, $\exists j . e_j \notin \chi_i$, should be equal no matter whether we base that belief on the original piece of metalevel evidence, before $e_q$ is taken out from $\chi_i$, or on a combination of the other two pieces of metalevel evidence $m_{\Delta\chi_i}(e_q \notin \chi_i)$ and $m_{\chi_i - \{e_q\}}(\exists j \neq q . e_j \notin (\chi_i - \{e_q\}))$, after $e_q$ is taken out from $\chi_i$, i.e.,

$$\text{Bel}_{\chi_i}(\exists j . e_j \notin \chi_i) = \text{Bel}_{\Delta\chi_i \oplus (\chi_i - \{e_q\})}(\exists j . e_j \notin \chi_i).$$

We get

$$m_{\Delta\chi_i}(e_q \notin \chi_i) = \frac{c_i - c_i^*}{1 - c_i^*},$$
$$m_{\Delta\chi_i}(\Theta) = \frac{1 - c_i}{1 - c_i^*}.$$

If $e_q$ then is brought into another subset $\chi_k$ its conflict will increase from $c_k$ to $c_k^*$. Thus, we will also have metalevel evidence regarding every other subset $\chi_k$,

$$\forall k \neq i . m_{\Delta\chi_k}(e_q \notin (\chi_k + \{e_q\})) = \frac{c_k^* - c_k}{1 - c_k},$$

$$\forall k \neq i . m_{\Delta\chi_k}(\Theta) = \frac{1 - c_k^*}{1 - c_k}.$$

### B.   Evidence From Domain Conflict Variations

Since a piece of evidence from domain conflict is evidence against the entire partitioning it is less specific than a piece of evidence from cluster conflict. We will interpret the domain conflict as evidence that there exists at least one piece of evidence that does not belong to any of the $n$ first subsets, $n = |\chi|$, or if this particular piece of evidence was placed in a subset by itself, as evidence that it belongs to some of the other $n - 1$ subsets. This would indicate that the number of subsets is incorrect.

We will now study any changes in the domain conflict when we take out a piece of evidence $e_q$ from subset $\chi_i$. When $|\chi_i| > 1$ we may not only put a piece of evidence $e_q$ that we have taken out from $\chi_i$ into another already existing subset, we may also put $e_q$ into a new subset $\chi_{n+1}$ by itself, assuming there are $n$ subsets, i.e., $\chi = \{\chi_1, ..., \chi_n\}$. This will change the domain conflict, $c_0$. Since the current partition minimizes the metaconflict function, we know that when the number of subsets increase we will get an increase in total conflict and Theorem 1 says that we will get a decrease in the nondomain part of the metaconflict function. Thus, we know that we must get an increase in the domain conflict. This increase in domain conflict is an indication that $e_q$ does not belong to an additional subset $\chi_{n+1}$.

Another way to receive a piece of evidence from the domain conflict is if $e_q$ is moved out from $\chi_i$ when $|\chi_i| = 1$. If $e_q$ is in a subset $\chi_i$ by itself and moved from $\chi_i$ to another already existing subset we may get either an increase or decrease in domain conflict. This is because both the total conflict and the nondomain part of the metaconflict function increases. Thus, we have two different situations. If the domain conflict decreases when we move $e_q$ out from $\chi_i$ this is interpreted as evidence that $e_q$ does not belongs to $\chi_i$, but if we receive an increase in domain conflict we will interpret this as evidence that $e_q$ does belong to $\chi_i$.

Let us analyze the case where we move $e_q$ from $\chi_i$ to $\chi_{n+1}$. When $e_q \in \chi_i$ and $|\chi_i| > 1$ we may move out $e_q$ from $\chi_i$ without changing the domain conflict, but we will get an increase in the domain conflict if we move $e_q$ to a subset by itself; $\chi_{n+1}$.

We get

$$m_{\Delta\chi}(e_q \notin \chi_{n+1}) = \frac{c_0^* - c_0}{1 - c_0}.$$

a piece of evidence indicating that $e_q$ does not belong to $\chi_{n+1}$.

Let us also study the situation when $e_q$ is in a subset by itself. If we take out $e_q$ from $\chi_i$, without moving it to any already existing subset, we have $n-1$ remaining subsets and get a decrease in the domain conflict. This decrease in domain conflict is interpreted as evidence that $e_q$ does not belong to $\chi_i$.

When $c_0 > c_0^*$ we have

$$m_{\Delta\chi}(e_q \notin \chi_i) = \frac{c_0 - c_0^*}{1 - c_0^*}.$$

A completely different situation occurs when $c_0 < c_0^*$. If we take out $e_q$ from $\chi_i$ as in the previous case, without moving it to any already existing subset, we have $n-1$ remaining subsets since $|\chi_i|$ was equal to one and get an increase in the domain conflict. This increase in domain conflict will be interpreted as evidence that $e_q$ belongs to $\chi_i$.

Thus, if $c_0 < c_0^*$ then

$$m_{\Delta\chi}(e_q \in \chi_i) = \frac{c_0}{c_0^*}$$

is derived as our piece of evidence.

## III. SPECIFYING EVIDENCE

We may now specify any original piece of evidence by combining all evidence from conflict variations regarding this particular piece of evidence. Then we may calculate the belief and plausibility for each subset that this particular piece of evidence belongs to the subset. The belief that it belongs to a subset will be zero, except when $e_q \in \chi_i$, $|\chi_i| = 1$ and $c_0 < c_0^*$, since every proposition regarding this piece of evidence then states that it does not belong to some subset.

In combining all pieces of evidence regarding an original piece of evidence we may receive support for a proposition stating that this piece of evidence does not belong to any of the subsets and cannot be put into a subset by itself. Since this is impossible, the statement is false and its support is the conflict in Dempster's rule. The statement that a piece of evidence does not belong anywhere implies that it is false. Thus, we may interpret the conflict as support for this piece of evidence being false.

Let us assume that a piece of evidence, $e_q$, is in $\chi_i$ and $|\chi_i| > 1$. When we combine all pieces of evidence regarding $e_q$ this results in a new basic probability assignment with

$$\forall \chi^* . m^*(e_q \notin (\ \vee \chi^*)) = \prod_{\chi_j \in \chi^*} m(e_q \notin \chi_j) \cdot \prod_{\chi_j \in (\chi - \chi^*)} [1 - m(e_q \notin \chi_j)]$$

where $\chi^* \in 2^\chi$, $\chi = \{\chi_1, ..., \chi_{n+1}\}$ and $\vee \chi^*$ is the disjunction of all elements in $\chi^*$.

From the new bpa we can calculate the conflict. The only statement that is false is the statement that $e_q \notin (\ \vee \chi\ )$, i.e., that $\forall j . e_q \in \chi_j$.

The conflict becomes

$$k = \frac{c_0^* - c_0}{1 - c_0} \cdot \prod_{j=1}^{n} \frac{c_j^* - c_j}{1 - c_j}.$$

When calculating belief and plausibility that $e_q$ belongs to some subset other than $\chi_{n+1}$ we have

$$\forall k \neq n+1 . \mathrm{Bel}(e_q \in \chi_k) = 0$$

and

$$\forall k \neq n+1 . \mathrm{Pls}(e_q \in \chi_k) = \frac{1 - \dfrac{c_k^* - c_k}{1 - c_k}}{1 - \dfrac{c_0^* - c_0}{1 - c_0} \cdot \prod\limits_{j=1}^{n} \dfrac{c_j^* - c_j}{1 - c_j}}$$

while for the subset $\chi_{n+1}$ we have

$$\mathrm{Bel}(e_q \in \chi_{n+1}) = 0$$

and

$$\mathrm{Pls}(e_q \in \chi_{n+1}) = \frac{1 - \dfrac{c_0^* - c_0}{1 - c_0}}{1 - \dfrac{c_0^* - c_0}{1 - c_0} \cdot \prod\limits_{j=1}^{n} \dfrac{c_j^* - c_j}{1 - c_j}}.$$

When $e_q \in \chi_i$, $|\chi_i| = 1$ and $c_0 > c_0^*$ the domain conflict variation appeared in the $i$th piece of evidence instead of the $n+1$th. With no evidence from a $n+1$th subset and domain conflict variation in the $i$th piece of evidence we have a slight change.

For subsets except $\chi_i$ we get

$$\forall k \neq i . \mathrm{Bel}(e_q \in \chi_k) = 0,$$

$$\forall k \neq i . \mathrm{Pls}(e_q \in \chi_k) = \frac{1 - \dfrac{c_k^* - c_k}{1 - c_k}}{1 - \dfrac{c_0 - c_0^*}{1 - c_0^*} \cdot \prod\limits_{\substack{j=1 \\ \neq i}}^{n} \dfrac{c_j^* - c_j}{1 - c_j}}$$

and for $\chi_i$

$$\mathrm{Bel}(e_q \in \chi_i) = 0,$$

$$\mathrm{Pls}(e_q \in \chi_i) = \frac{1 - \dfrac{c_0^* - c_0}{1 - c_0}}{1 - \dfrac{c_0^* - c_0}{1 - c_0} \cdot \prod\limits_{\substack{j=1 \\ \neq i}}^{n} \dfrac{c_j^* - c_j}{1 - c_j}}.$$

When $e_q \in \chi_i$, $|\chi_i| = 1$ and $c_0 < c_0^*$ we received an increase in domain conflict when $e_q$ was moved out from $\chi_i$. This introduced a new type of evidence supporting that $e_q$ belongs to $\chi_i$. Since we did not have a piece of evidence indicating that $e_q$ did not belong to $\chi_i$ we will never have any support for the impossible statement that $e_q$ does not belong anywhere. Thus, we will always get a zero conflict when combining these pieces of evidence.

When calculating belief and plausibility for any subset other than $\chi_i$ we get

$$\forall k \neq i.\mathrm{Bel}(e_q \in \chi_k) = 0$$

and

$$\forall k \neq i.\mathrm{Pls}(e_q \in \chi_k) = \left(1 - \frac{c_0}{c_0^*}\right) \cdot \left(1 - \frac{c_k^* - c_k}{1 - c_k}\right)$$

and for $\chi_i$:

$$\mathrm{Bel}(e_q \in \chi_i) = \frac{c_0}{c_0^*} + \left(1 - \frac{c_0}{c_0^*}\right) \cdot \prod_{\substack{j = 1 \\ \neq i}}^{n} \frac{c_j^* - c_j}{1 - c_j}$$

and

$$\mathrm{Pls}(e_q \in \chi_i) = 1$$

because of the lack of evidence against that $e_q$ belongs to the $i$th subset.

## IV.  HANDLING THE FALSITY OF EVIDENCE

In Section III we received support $k$ for the statement that a piece of evidence $e_q$ did not belong to any of the subsets. Since this is impossible the statement implies to a degree $k$ that $e_q$ is a false piece of evidence. If a piece of evidence is known to be false we would disregard it completely, and when we have no indication as to the possible falsity of it we would take no additional action.

We would then like to pay less and less regard to a piece of evidence the higher the degree is that it is false, pay no attention to it when it is certainly false, and leave it unchanged when there is no indication as to its falsity. This can be done by using the discounting operation introduced by Lowrance et al.[10] The discounting operation was introduced to handle the case when the source of some piece of evidence is lacking in credibility. The credibility of the source, $\alpha$, also became the credibility of the piece of evidence. The situation was handled by discounting each supported proposition other than $\Theta$ with the credibility $\alpha$ and by adding the discounted mass to $\Theta$;

$$m^{\%}(A_j) = \begin{cases} \alpha \cdot m(A_j), & A_j \neq \Theta \\ 1 - \alpha + \alpha \cdot m(\Theta), & A_j = \Theta \end{cases}.$$

We will use the same discounting operation in this case when there is a direct indication for each separate piece of evidence regardless of which source produced it. We will view the support of the false statement that a piece of evidence $e_q$ does not belong to any subset and cannot be put in a subset by itself, i.e., the conflict in Dempster's rule when combining all pieces of evidence regarding $e_q$, as identical to one minus the credibility of the evidence;

$$\alpha \triangleq 1 - m^*(e_q \notin (\quad \vee \chi)) = 1 - k.$$

Thus, a piece of evidence is discounted in relation to its degree of falsity.

It is obvious that the credibility used to discount a piece of evidence depends on the piece of evidence itself. This should be no problem since the credibility originates from a piece evidence at a "higher" level that depends on $e_q$ but will never be combined with $e_q$. Instead, it is used to discount $e_q$. Obviously, any discounting directed towards individual pieces of evidence and not all pieces of evidence from a particular source will depend on the piece of evidence itself.

We should note that we must not repartition the set of all pieces of evidence after the first discounting in order to receive new credibilities and perform a second discounting. The two pieces of evidence from which the two credibilities are originating would not be independent. Thus, making a second discounting of any piece of evidence would violate the independence assumption of Dempster's rule since a double discounting corresponds to combining the two nonindependent pieces of evidence concerning the falsity of $e_q$ and discounting $e_q$ with the credibility of $e_q$ derived from the combination;

$$\alpha_{12} = 1 - k_{12} = 1 - [1 - (1 - k_1) \cdot (1 - k_2)] = (1 - k_1) \cdot (1 - k_2) = \alpha_1 \cdot \alpha_2$$

where $\alpha_1$ and $\alpha_2$ are the two credibilities of the first and second discounting and $\alpha_{12}$ is the credibility derived from the combination of both corresponding pieces of evidence, and $k_1$ and $k_2$ are the two degrees of falsity of the first and second partitioning and $k_{12}$ is the degree of falsity in the combination of the two pieces of evidence.

In fact, we should never repartition evidence after discounting, regardless of whether we plan to perform a second discounting or not. The discounting operation not only puts the evidence in order for continuing reasoning processes regarding the different events, but it also smooths out the conflicting differences between different pieces of evidence which is the very basis of the conflict minimizing process when the set of all pieces of evidence are partitioned into subsets. Since the discounting smooths out the differences between pieces of evidence that do not belong to the same event, a repartitioning would only increase the risk that pieces of evidence referring to different events would be partitioned into the same subset. Thus, we should never repartition the set of all pieces of evidence after discounting evidence for falsity.

The evidence we specified in Section III may now be discounted to its degree of credibility:

evidence $q$:
   proposition:
      action part: $A_f, A_g, ..., A_h$
      event part: $[\text{Bel}(e_q \in \chi_i), \text{Pls}(e_q \in \chi_i)] / E_i, \ [0, \text{Pls}(e_q \in \chi_j)] / E_j, ...,$
                  $[0, \text{Pls}(e_q \in \chi_k)] / E_k$

$m(A_f) = \alpha \cdot p_f$
$m(A_g) = \alpha \cdot p_g$
...
$m(A_h) = \alpha \cdot p_h$
$m(\Theta) = 1 - \alpha \cdot p_f - \alpha \cdot p_g - ... - \alpha \cdot p_h$

where $\alpha$ is the degree of credibility and $1 - \alpha$ is the degree of falsity of $e_q$.

## V.  FINDING USABLE EVIDENCE

Obviously, the next question to put is: Will our now specified and dis-counted piece of evidence be of use in a subsequent reasoning process concern-ing a particular event? If this piece of evidence can only belong to one subset then it is also usable in a subsequent reasoning process for that subset. Whether this is the case or not will be determined by the now specified event part. If the piece of evidence will be useful in the reasoning process as well is another question. That depends only on the action part of the piece of evidence.

If a piece of evidence can belong to more than one subset it will clearly be uncertain if it belongs to our subset in question if indeed that is possible at all. We must find a measure of this uncertainty — a credibility that it belongs to the subset. Before using a piece of evidence in the reasoning process concern-ing our subset, we would like to calculate the credibility that it belongs to the subset in question and then discount it by its credibility. Obviously, a piece of evidence that cannot possibly belong to a subset $\chi_i$ should be discounted entirely in the subsequent reasoning process for that subset, while a piece of evidence which cannot possibly belong to any other subset $\chi_j$ and is without any support whatsoever against $\chi_i$ should not be discounted at all when used in the reasoning process for $\chi_i$. Thus, the degree to which a piece of evidence can belong to a subset and no other subset corresponds to the importance it should be allowed to play in that subset.

In order to find the credibility of a piece of evidence in the reasoning process for some subset we must measure the uncertainty in the newly specified event part. Measures of uncertainty in a single piece of evidence are usually measures of entropy. An especially useful kind of such measure is the measure of average total uncertainty.[11,12] This is a measure of entropy that measures both scattering and nonspecificity of evidence:

$$H(m) = \sum_{A \in \Theta} m(A) \cdot \text{Log}_2(|A|) - \sum_{A \in \Theta} m(A) \cdot \text{Log}_2(m(A)).$$

However, the average total uncertainty in which event a piece of evidence $e_q$ might be referring to is not exactly our concern. This measure applied to the new basic probability assignment resulting from the fusion of all derived pieces of evidence regarding to which subset $e_q$ can belong (Sec. III-A) would give us an indication of how usable this piece of evidence is in total towards all subsets. What we want is a high plausibility for the most likely subset, i.e., little support against that the piece of evidence belongs to the subset. This is equivalent with preferring a minimal entropy $H(m)$, but how the remainder of the support is scattered among the other focal elements is of little concern to us. Actually, if we are to express some preference regarding the remainder of the support we would choose some uniform scattering among the other focal elements, i.e., preferring as low as possible a plausibility for the second most likely subset. This is not equivalent with preferring a minimal entropy. When it comes to the specificity of the support against different subsets, we prefer such a support to be specific when it concerns other subsets and most preferably gives these subsets a low plausibility, and to have some nonspecificity when it concerns the most preferable subset giving it a plausibility as high as possible. Thus, our overall preference is not consistent with a minimal entropy.

We might be able to find some entropy-like measure of entropy difference between two parts of a piece of evidence that could be maximized. But rather than going this route we will make some simple observations (axioms):

- If the plausibility that $e_q$ belongs to some subset is zero we should discount it entirely.
- If the plausibility of a subset is one and the plausibility of all other subsets are zero we should not discount $e_q$ at all when used in this subset.
- If the plausibility of a subset is $\alpha$, while the belief is zero, and the plausibility of all other subsets are zero we should discount $e_q$ to a credibility of $\alpha$.
- If the plausibility of $n$ different subsets are all one and the plausibility of all other subsets are zero we should discount $e_q$ to a credibility of $1/n$.
- The credibility of $e_q$ when used in a subset is greater or equal to the belief of the subset.

A function that satisfies these observations is the plausibility of the subset weighted by the portion of the plausibility for all subsets that this subset has received and by the portion of the still uncommitted belief.

The credibility $\alpha_j$ of $e_q$ when $e_q$ is used in $\chi_j$ can then be calculated as

$$\alpha_j = [\,1 - \mathrm{Bel}(e_q \in \chi_i)\,] \cdot \frac{[\mathrm{Pls}(e_q \in \chi_j)]^2}{\sum_k \mathrm{Pls}(e_q \in \chi_k)}, j \neq i,$$

$$\alpha_i = \mathrm{Bel}(e_q \in \chi_i) + [\,1 - \mathrm{Bel}(e_q \in \chi_i)\,] \cdot \frac{[\mathrm{Pls}(e_q \in \chi_i)]^2}{\sum_k \mathrm{Pls}(e_q \in \chi_k)}.$$

Here, $\mathrm{Bel}(e_q \in \chi_i)$ is equal to zero except when $e_q \in \chi_i$, $|\chi_i| = 1$ and $c_0 < c_0^*$.

The discounting we make of $e_q$ should not be confused with the discounting we made in Section IV. That discounting was made "on principle" due to the derived evidence proposing to some degree that $e_q$ was false. The discounting we are making here, however, is merely a technical necessity in order to be able to use the evidence when we as users force an absolute specificity upon the event part of a piece of evidence by placing it in one of the subsets.

After discounting each piece of evidence to its new credibility in a particular subset the subsequent reasoning process could begin. Note that a piece of evidence could be used in several different subsets with an appropriate discounting, i.e., for a particular subset every piece of evidence that belongs to the subset with a plausibility above zero could be used in the reasoning process within that subset. Also, a piece of evidence whose original event part only indicated one possible event, say $E_j$, and which now has a plausibility of one for some subsets $\chi_i$ might still have a credibility below one for $\chi_i$. This should come as no surprise since our piece of evidence might not have been completely certain, i.e., leaving some mass on $\Theta$. Since the mass on $\Theta$ supports any event, our piece of evidence is not completely certain regarding which event it is referring to, giving us a credibility below one for $\chi_i$. Also, even if it was completely certain, then $\chi_i$, whose meaning is determined by all pieces of evidence it contains, might not for certain be representing $E_j$.

## VI. FINDING THE EVENT OF A SUBSET

When we begin our subsequent reasoning process in each subset, it will naturally be of vital importance to know to which event the subset is referring. This information is obtainable when the pieces of evidence in the subset have been combined. After the combination, each focal element of the final bpa will in addition to supporting some proposition regarding an action also be referring to one or more events where the proposed action may have taken place. We could simply sum up the support in favor of each event, calculate the plausibility of it, and then form our opinion regarding which event the subset is referring to based on this result. However, this may cause a problem. It would certainly be possible that more than one subset has one and the same event as its most likely event. This situation can be avoided if we bring the problem to the metalevel where we simultaneously reason about all subsets, i.e., which subsets are referring to which events. In this analysis we use our domain knowledge stating that no more than one subset may be referring to an event. From each subset we have a piece of evidence indicating which events it might be referring to. This piece of evidence is directly derivable from the final bpa resulting from the combination of all pieces of evidence in the subset. We simply remove the information about action from each focal element in the final bpa while leaving the information about event unchanged. This may leave us with two or more focal elements supporting the same event or disjunction of events. The support for these focal elements are summed up and the focal elements are represented only once. That is, we receive a new piece of evidence at the metalevel originating from the subset that is not paying any attention to actions but paying the same attention to events as the final bpa resulting from the combination of all pieces of evidence within this subset. Thus, we have the following bpa of the piece of evidence originating from $\chi_i$:

$$\forall E.m_{\chi_i}((\ \lor E)/\chi_i) = \sum_{Event\ part\ of\ A\ is\ E} m_{\chi_i}(A)$$

where $E \in 2^{\{E_j\}}$. Here, of course, $\lor \{E_j\}$ is $\Theta$.

Combining all bpa's from all different subsets with the restriction that any intersection in the combination yielding $E_k/\chi_i \land E_k/\chi_j$ is false eliminates the possible problem of having an event simultaneously assigned to two or more different subsets. This method has a much higher chance to give a clearly preferable answer regarding which event are represented by which subsets, than that of only viewing the pieces of evidence within a subset when trying to determine its event.

## VII.    AN EXAMPLE

Let us return to the problem of two possible burglaries described in our first article.[4] We will now reexamine this problem in view of the results of Sections II to VI. Finally, we make a comparison between an overconfident approach of only partitioning the pieces of evidence by minimizing the metaconflict function before we begin the reasoning process separately in each subset, and the refined approach of discounting for falsity and uncertainty in affiliation proposed in this article.

### A.    A Refined Analysis of the Bakers' Shops Burglaries

In this example we had evidence weakly specified in the sense that it is uncertain to which possible burglary the propositions are referring. We will try to specify these pieces of evidence by studying cluster conflict variations when one piece of evidence is moved from its subset to another subset or put into a new subset by itself. The problem we were facing was described as follows:[4]

Assume that a baker's shop at One Baker Street has been burglarized, event 1. Let there also be some indication that a baker's shop across the street, at Two Baker Street, might have been burglarized, although no burglary has been reported, event 2. An experienced investigator estimates that a burglary has taken place at Two Baker Street with a probability of 0.4. We have received the following pieces of evidence. A credible witness reports that "a brown-haired man who is not an employee at the baker's shop committed the burglary at One Baker Street," evidence 1. An anonymous witness, not being aware that there might be two burglaries, has reported "a brown-haired man who works at the baker's shop committed the burglary at Baker Street," evidence 2. Thirdly, a witness reports having seen "a suspicious-looking red-haired man in the baker's shop at Two Baker Street," evidence 3. Finally, we have a fourth witness, this witness, also anonymous and not being aware of the possibility of two burglaries, reporting that the burglar at the Baker Street baker's shop was a brown-haired man. That is, for example:

evidence 1:                           evidence 2:
   proposition:                        proposition:
      action part: *BO*                 action part: *BI*

|                          |                          |
|--------------------------|--------------------------|
| event part: $E_1$:       | event part: $E_1, E_2$   |
| $m(BO) = 0.8$            | $m(BI) = 0.7$           |
| $m(\Theta) = 0.2$        | $m(\Theta) = 0.3$       |

evidence 3:
   proposition:
      action part: $R$

evidence 4:
   proposition:
      action part: $B$

      event part: $E_2$
$m(R) = 0.6$
$m(\Theta) = 0.4$

      event part: $E_1, E_2$
$m(B) = 0.5$
$m(\Theta) = 0.5$

domain probability distribution:

$$m(E_i) = \begin{cases} 0.6, & i = 1 \\ 0.4, & i = 2 \\ 0, & i \neq 1, 2 \end{cases}.$$

All pieces of evidence where originally put into one subset, $\chi_1$. By minimizing the metaconflict function it was found best to partition the pieces of evidence into two subsets. The minimum of the metaconflict function was found when evidence one and four were moved from $\chi_1$ into $\chi_2$ while evidence two and three remained in $\chi_1$. From the event parts of the pieces of evidence we were able to conclude that $\chi_1$ corresponded to event 2 and $\chi_2$ corresponded to event 1.

Let us now study the cluster conflict variations. The conflict in $\chi_1$ was $c_1 = 0.42$, in $\chi_2$ it was $c_2 = 0$, with a domain conflict of $c_0 = 0.6$. If $e_1$ now in $\chi_2$ is moved out from $\chi_2$ the conflict will drop to zero, $c_2^* = 0$. If $e_1$ is then moved into $\chi_1$ its conflict increases to $c_1^* = 0.788$, and if $e_1$ is put into a subset by itself, $\chi_3$, we will have a domain conflict of one, $c_0^* = 1$.

Thus, by the formulas of Section II-C we get

$$m(e_1 \notin \chi_1) = \frac{c_1^* - c_1}{1 - c_1} = 0.634, \qquad m(e_1 \notin \chi_2) = \frac{c_2 - c_2^*}{1 - c_2^*} = 0$$

and

$$m(e_1 \notin \chi_3) = \frac{c_0^* - c_0}{1 - c_0} = 1.$$

From these pieces of evidence we will calculate the plausibility for each subset that $e_1$ belongs to the subset:

$$\text{Pls}(e_1 \in \chi_1) = \frac{1 - m(e_1 \notin \chi_1)}{1 - m(e_1 \notin \chi_1) \cdot m(e_1 \notin \chi_2) \cdot m(e_1 \notin \chi_3)} = 1 - 0.634 = 0.366,$$

$$\text{Pls}(e_1 \in \chi_2) = \frac{1 - m(e_1 \notin \chi_2)}{1 - m(e_1 \notin \chi_1) \cdot m(e_1 \notin \chi_2) \cdot m(e_1 \notin \chi_3)} = 1,$$

$$\text{Pls}(e_1 \in \chi_3) = \frac{1 - m(e_1 \notin \chi_3)}{1 - m(e_1 \notin \chi_1) \cdot m(e_1 \notin \chi_2) \cdot m(e_1 \notin \chi_3)} = 0.$$

We do the same for the other three pieces of evidence:

$$m(e_2 \notin \chi_i) = \begin{cases} 0.42, i = 1 \\ 0.56, i = 2, \\ 1, i = 3 \end{cases} \qquad m(e_3 \notin \chi_i) = \begin{cases} 0.42, i = 1 \\ 0.54, i = 2, \\ 1, i = 3 \end{cases}$$

$$m(e_4 \notin \chi_i) = \begin{cases} 0.155, i = 1 \\ 0, i = 2 \\ 1, i = 3 \end{cases}$$

and calculate the plausibilities

$$\text{Pls}(e_2 \in \chi_i) = \begin{cases} 0.758, i = 1 \\ 0.575, i = 2, \\ 0, i = 3 \end{cases} \qquad \text{Pls}(e_3 \in \chi_i) = \begin{cases} 0.750, i = 1 \\ 0.595, i = 2, \\ 0, i = 3 \end{cases}$$

$$\text{Pls}(e_4 \in \chi_i) = \begin{cases} 0.845, i = 1 \\ 1, i = 2 \\ 0, i = 3 \end{cases}.$$

Thus, the four pieces of evidence are specified as:

evidence 1:
  proposition:
    action part: $BO$
    event part:
    $[0, 0.366]/\chi_1, [0, 1]/\chi_2$
  $m(BO) = 0.8$
  $m(\Theta) = 0.2$

evidence 2:
  proposition:
    action part: $BI$
    event part:
    $[0, 0.758]/\chi_1, [0, 0.575]/\chi_2$
  $m(BI) = 0.7$
  $m(\Theta) = 0.3$

evidence 3:
  proposition:
    action part: $R$
    event part:
    $[0, 0.750]/\chi_1, [0, 0.595]/\chi_2$
  $m(R) = 0.6$
  $m(\Theta) = 0.4$

evidence 4:
  proposition:
    action part: $B$
    event part:
    $[0, 0.845]/\chi_1, [0, 1]/\chi_2$
  $m(B) = 0.5$
  $m(\Theta) = 0.5$

Thus, it seems pretty certain that $e_1$ belongs to $\chi_2$ while the other three pieces of evidence are more uncertain in which subset they belong to, i.e., more nonspecific in which event they are referring to. Especially $e_4$ is not specific. It could almost belong to either subset.

When we combined the pieces of evidence regarding where a particular piece of evidence might belong, we received a conflict for $e_2$ and $e_3$ but not for $e_1$ and $e_4$. Thus, there is no indication that $e_1$ and $e_4$ might be false. For the second and third evidence we got a conflict of 0.2352 and 0.2268, respectively.

This is their degrees of falsity. We should then discount $e_2$ and $e_3$ to their respective degrees of credibility, i.e., 0.7648 and 0.7732:

evidence 1:
    proposition:
        action part: *BO*
        event part:
           $[0, 0.366]/\chi_1, [0, 1]/\chi_2$
    $m(BO) = 0.8$
    $m(\Theta) = 0.2$

evidence 2:
    proposition:
        action part: *BI*
        event part:
           $[0, 0.758]/\chi_1, [0, 0.575]/\chi_2$
    $m(BI) = 0.5354$
    $m(\Theta) = 0.4646$

evidence 3:
    proposition:
        action part: *R*
        event part:
           $[0, 0.750]/\chi_1, [0, 0.595]/\chi_2$
    $m(R) = 0.4639$
    $m(\Theta) = 0.5361$

evidence 4:
    proposition:
        action part: *B*
        event part:
           $[0, 0.845]/\chi_1, [0, 1]/\chi_2$
    $m(B) = 0.5$
    $m(\Theta) = 0.5$

The discounting of $e_2$ and $e_3$ due to their fairly high degree of falsity will reduce the impact of these two pieces of evidence in a subsequent reasoning process regarding the two different events.

Before we finally start the reasoning process in $\chi_1$ and $\chi_2$ we should once again discount the pieces of evidence. This time we make an individual discounting for each subset and piece of evidence according to how credible it is that the piece of evidence belongs to the subset. The credibility that $e_1$ belongs to $\chi_1$ is

$$\alpha_1 = \frac{(\mathrm{Pls}(e_1 \in \chi_1))^2}{\displaystyle\sum_{j=1}^{2} \mathrm{Pls}(e_1 \in \chi_j)} = \frac{0.366^2}{0.366 + 1} = 0.0981$$

and that $e_1$ belongs to $\chi_2$

$$\alpha_2 = \frac{(\mathrm{Pls}(e_1 \in \chi_2))^2}{\displaystyle\sum_{j=1}^{2} \mathrm{Pls}(e_1 \in \chi_j)} = \frac{1}{0.366 + 1} = 0.7321.$$

For the other three pieces of evidence we get: $e_2$: $\alpha_1 = 0.4310$, $\alpha_2 = 0.2480$, $e_3$: $\alpha_1 = 0.4182$, $\alpha_2 = 0.2632$, and for $e_4$: $\alpha_1 = 0.3870$, $\alpha_2 = 0.5420$. Discounting the four pieces of evidence to their credibility of belonging to $\chi_1$ and $\chi_2$, respectively, yields:

evidence 1:
    proposition:
        action part: *BO*
        event part:
           $[0, 0.366]/\chi_1, [0, 1]/\chi_2$
    $m(BO) = 0.0784/\chi_1, 0.5856/\chi_2$
    $m(\Theta) = 0.9216/\chi_1, 0.4144/\chi_2$

evidence 2:
    proposition:
        action part: *BI*
        event part:
           $[0, 0.758]/\chi_1, [0, 0.575]/\chi_2$
    $m(BI) = 0.2308/\chi_1, 0.1328/\chi_2$
    $m(\Theta) = 0.7692/\chi_1, 0.8672/\chi_2$

evidence 3:
  proposition:
    action part: $R$
    event part:
      $[0, 0.750]/\chi_1, [0, 0.595]/\chi_2$
$m(R) = 0.1940/\chi_1, 0.1221/\chi_2$
$m(\Theta) = 0.8060/\chi_1, 0.8779/\chi_2$

evidence 4:
  proposition:
    action part: $B$
    event part:
      $[0, 0.9216]/\chi_1, [0, 0.4144]/\chi_2$
$m(B) = 0.1935/\chi_1, 0.2710/\chi_2$
$m(\Theta) = 0.8065/\chi_1, 0.7290/\chi_2$

Combining these four pieces of evidence with Dempster's rule results in the following final basic probability assignment:

$$
\begin{aligned}
m^*_{1 \oplus 2 \oplus 3 \oplus 4}(BO \wedge E_1) &= \frac{1}{1-k} \cdot m_1(BO \wedge E_1) \\
&\quad \cdot [1 - m_2(BI \wedge (E_1 \vee E_2))] \cdot [1 - m_3(R \wedge E_2)] \\
&= 0.0539/\chi_1, 0.5298/\chi_2,
\end{aligned}
$$

$$
\begin{aligned}
m^*_{1 \oplus 2 \oplus 3 \oplus 4}(BI \wedge (E_1 \vee E_2)) &= \frac{1}{1-k} \cdot [1 - m_1(BO \wedge E_1)] \\
&\quad \cdot m_2(BI \wedge (E_1 \vee E_2)) \cdot [1 - m_3(R \wedge E_2)] \\
&= 0.1900/\chi_1, 0.0574/\chi_2,
\end{aligned}
$$

$$
\begin{aligned}
m^*_{1 \oplus 2 \oplus 3 \oplus 4}(B \wedge (E_1 \vee E_2)) &= \frac{1}{1-k} \cdot [1 - m_1(BO \wedge E_1)] \\
&\quad \cdot [1 - m_2(BI \wedge (E_1 \vee E_2))] \cdot [1 - m_3(R \wedge E_2)] \\
&\quad \cdot m_4(B \wedge (E_1 \vee E_2)) = 0.1225/\chi_1, 0.1016/\chi_2,
\end{aligned}
$$

$$
\begin{aligned}
m^*_{1 \oplus 2 \oplus 3 \oplus 4}(R \wedge E_2) &= \frac{1}{1-k} \cdot [1 - m_1(BO \wedge E_1)] \\
&\quad \cdot [1 - m_2(BI \wedge (E_1 \vee E_2))] \cdot m_3(R \wedge E_2) \\
&\quad \cdot [1 - m_4(B \wedge (E_1 \vee E_2))] = 0.1229/\chi_1, 0.0380/\chi_2,
\end{aligned}
$$

$$
\begin{aligned}
m^*_{1 \oplus 2 \oplus 3 \oplus 4}(\Theta) &= \frac{1}{1-k} \cdot [1 - m_1(BO \wedge E_1)] \\
&\quad \cdot [1 - m_2(BI \wedge (E_1 \vee E_2))] \\
&\quad \cdot [1 - m_3(R \wedge E_2)] \cdot [1 - m_4(B \wedge (E_1 \vee E_2))] \\
&= 0.5107/\chi_1, 0.2732/\chi_2
\end{aligned}
$$

where $k$ is the conflict in Dempster's rule;

$$
\begin{aligned}
k &= m_3(R \wedge E_2) \cdot (1 - [1 - m_1(BO \wedge E_1)] \cdot [1 - m_2(BI \wedge (E_1 \vee E_2))] \\
&\quad \cdot [1 - m_4(B \wedge (E_1 \vee E_2))]) + [1 - m_3(R \wedge E_2)] \cdot m_1(BO \wedge E_1) \\
&\quad \cdot m_2(BI \wedge (E_1 \vee E_2)) = 0.0977/\chi_1, 0.1584/\chi_2.
\end{aligned}
$$

Finally, this gives us the following evidential intervals:

$$[\text{Bel}(BO), \text{Pls}(BO)] = [0.0539, 0.6871]/\chi_1, [0.5298, 0.9046]/\chi_2,$$
$$[\text{Bel}(BI), \text{Pls}(BI)] = [0.1900, 0.8232]/\chi_1, [0.0574, 0.4322]/\chi_2,$$
$$[\text{Bel}(B), \text{Pls}(B)] = [0.3664, 0.8771]/\chi_1, [0.6888, 0.9620]/\chi_2,$$
$$[\text{Bel}(R), \text{Pls}(R)] = [0.1229, 0.6336]/\chi_1, [0.0380, 0.3112]/\chi_2,$$
$$[\text{Bel}(I), \text{Pls}(I)] = [0.1900, 0.9461]/\chi_1, [0.0574, 0.4702]/\chi_2,$$
$$[\text{Bel}(O), \text{Pls}(O)] = [0.0539, 0.8100]/\chi_1, [0.5298, 0.9426]/\chi_2,$$
$$[\text{Bel}(E_1), \text{Pls}(E_1)] = [0.0539, 0.8771]/\chi_1, [0.5298, 0.9620]/\chi_2,$$
$$[\text{Bel}(E_2), \text{Pls}(E_2)] = [0.1229, 0.9461]/\chi_1, [0.0380, 0.4702]/\chi_2.$$

From the intervals regarding which event the subsets are referring to it is somewhat uncertain whether $\chi_1$ is referring to $E_1$ or $E_2$. However, $\chi_2$ clearly refers to $E_1$.

Let us bring the problem to the metalevel together with our domain knowledge that the two subsets must be referring to different events. We create two new but very similar basic probability assignments as follows:

$$m_{\chi_1}(E_1/\chi_1) = 0.0539,$$
$$m_{\chi_1}(E_2/\chi_1) = 0.1229,$$
$$m_{\chi_1}(\Theta) = 1 - m_{\chi_1}(E_1/\chi_1) + m_{\chi_1}(E_2/\chi_1) = 0.8232$$

and

$$m_{\chi_2}(E_1/\chi_2) = 0.0539,$$
$$m_{\chi_2}(E_2/\chi_2) = 0.1229,$$
$$m_{\chi_2}(\Theta) = 1 - m_{\chi_2}(E_1/\chi_2) + m_{\chi_2}(E_2/\chi_2) = 0.8232$$

Combining these on the metalevel yields

$$m_{\chi_1 \oplus \chi_2}(E_1/\chi_1 \wedge E_2/\chi_2) = 0.0699,$$
$$m_{\chi_1 \oplus \chi_2}(E_2/\chi_1 \wedge E_1/\chi_2) = 0.6840,$$
$$m_{\chi_1 \oplus \chi_2}(\Theta) = 0.2462$$

with evidential intervals

$$[\text{Bel}(E_1/\chi_1 \wedge E_2/\chi_2), \text{Pls}(E_1/\chi_1 \wedge E_2/\chi_2)] = [0.0699, 0.3160]$$

and

$$[\text{Bel}(E_2/\chi_1 \wedge E_1/\chi_2), \text{Pls}(E_2/\chi_1 \wedge E_1/\chi_2)] = [0.6840, 0.9301].$$

This makes it perfectly clear that $\chi_1$ refers to $E_2$ while $\chi_2$ refers to $E_1$.

We see in conclusion that at $\chi_1$, i.e., event 2, there is some support for the burglar being brown-haired although it is certainly plausible, although less likely, he was actually red-haired. We have an even slighter indication that this might be an inside job but it is also possible that the burglar was an outsider. In general the evidence regarding event 1 is pretty inconclusive. However, the picture is much clearer at $\chi_2$, i.e., event 1. It is quite likely that the burglar at event 1 was a brown-haired outsider.

### B.   A Comparison Between an Overconfident and a Refined Analysis of the Bakers' Shops Burglaries

When we partitioned the four pieces of evidence, $e_2$ and $e_3$ ended up in $\chi_1$ while $e_1$ and $e_4$ ended up in $\chi_2$. This was the partitioning that minimized the metaconflict and thus the most probable partition. However, it said nothing about the probability for some piece of evidence that it belong to the subset where it was placed, and nothing about how much more probable this subset was to other subsets. It only said that this was the most probable subset of all. Thus, a piece of evidence might end up in some subset that was only marginally better than some other. This somewhat overconfident approach might then falsely indicate a certainty in the subsequent reasoning process within each subset that does not really exist. This false certainty is due to the restriction of not, to any degree, using pieces of evidence that ended up in other subsets by the partitioning. In contrast, the refined approach uses all pieces of evidence that could possibly belong to a subset in the reasoning process for that subset, although they are discounted to the credibility that they belong to the subset. This approach eliminates the problem of false certainty imposed by the partitioning as seen in the following comparison of the two approaches applied to the bakers' shops burglary problem.

As a comparison between the two approaches there is not much to say about the conclusions drawn in $\chi_2$. Whatever was concluded earlier is also concluded in the refined approach. That is, our burglar is a brown-haired outsider. The only real difference seems to be a somewhat higher plausibility for unlikely red-haired and insider alternatives together with a lower support for the preferred brown-haired and outsider alternatives due mainly to the possibility that $e_2$ or $e_3$ placed in $\chi_1$ by the partitioning might belong to the subset.

As before, the situation is not so clear at $\chi_1$. In general we find that evidential intervals have opened up in the refined approach. This is due to the discounting of evidence. In the refined approach we see an especially large drop in support for alternatives supported by pieces of evidence that belong to $\chi_1$ in the overconfident approach, brown-haired insider and red-haired, and also a large increase in plausibility for alternatives supported by pieces of evidence that belong to $\chi_2$, brown-haired and brown-haired outsider. This is due to the possibility that the pieces of evidence that belong to $\chi_2$ in the overconfident approach actually has a possibility of belonging instead to $\chi_1$, and vice versa. If we consider the three alternatives brown-haired insider, brown-haired and insider in $\chi_1$ they all had a support of 0.483 and a plausibility of 0.69, 0.69 and 1, respectively, in the overconfident approach. In the refined approach there is only a small drop in support for brown-haired to 0.36 but a much larger drop to 0.19 for insider and brown-haired insider. This is due to the possibility that $e_1$ supporting brown-haired outsider belongs to $\chi_1$. This might not be very plausible but if it was the case it would have a large impact since $e_1$ is strongly supportive of brown-haired outsider. Thus, in the overconfident approach we might have falsely concluded that the burglar was a brown-haired insider while it actually, as shown in the refined approach, is much more of an open question whether the probably brown-haired burglar was an insider or not.

## VIII.  CONCLUSIONS

In this article we have extended the methodology to partition nonspecific evidence developed in our previous article[4] to a methodology for specifying nonspecific evidence. This is in itself clearly an important extension in analysis, considering that a piece of evidence will now in a subsequent reasoning process be handled similarly by different subsets if these are approximately equally plausible, whereas before the most plausible subset would take the piece of evidence as certainly belonging to the subset while the other subsets would never consider it in their reasoning processes. In addition, two facts will facilitate the reasoning process. First, the specification process in the extended methodology will besides specifying all pieces of evidence also give a degree of falsity and a degree of credibility in affiliation for each piece of evidence. Secondly, the methodology can iteratively receive its pieces of evidence piece by piece. Together, these facts indicate that it should be possible to develop methods for disregarding immediately upon receipt false pieces of evidence as well as methods for focusing attention upon useful pieces of evidence based on their maximum degree of credibility.

The work described in this article has been further extended[13] to find a posterior probability distribution regarding the number of subsets (also in my recent Ph.D. thesis[14, 15]). For this we used the idea that each piece of evidence in a subset supports the existence of that subset to the degree that this piece of evidence supports anything at all. We can then derive a bpa that is concerned with the question of how many subsets we have. That bpa is combined with a given prior domain probability distribution in order to obtain the sought-after posterior domain distribution.

I would like to thank Stefan Arnborg, Ulla Bergsten, and Per Svensson for their helpful comments regarding this article.

## APPENDIX I:   EVIDENCE ABOUT EVIDENCE

### A.  Evidence From Cluster Conflict Variations

A conflict in a subset can be interpreted as a piece of metalevel evidence that there is at least one piece of evidence that does not belong to the subset. Thus, we can refine the basic probability assignment from subset $\chi_i$ assigning support to a proposition against the partitioning:

$$m_{\chi_i}(\neg AdP) = c_i,$$
$$m_{\chi_i}(\Theta) = 1 - c_i$$

to

$$m_{\chi_i}(\exists j . e_j \notin \chi_i) = c_i,$$
$$m_{\chi_i}(\Theta) = 1 - c_i.$$

Let us observe one piece of evidence $e_q$ in $\chi_i$. If $e_q$ is taken out from $\chi_i$ the conflict $c_i$ in $\chi_i$ decreases to $c_i^*$. This decrease in conflict $c_i - c_i^*$ can be interpreted as follows: there exists some metalevel evidence indicating that $e_q$ does not belong to $\chi_i$,

$$m_{\Delta\chi_i}(e_q \notin \chi_i),$$
$$m_{\Delta\chi_i}(\Theta),$$

and the remainder of the conflict $c_i^*$ is metalevel evidence that there is at least one other piece of evidence $e_j$, $j \neq q$, that does not belong to $\chi_i - \{e_q\}$,

$$m_{\chi_i - \{e_q\}}(\exists j \neq q . e_j \notin (\chi_i - \{e_q\})) = c_i^*,$$
$$m_{\chi_i - \{e_q\}}(\Theta) = 1 - c_i^*.$$

We will derive the basic probability number of $e_q \notin \chi_i$, $m_{\Delta\chi_i}(e_q \notin \chi_i)$, by stating that the belief in the proposition that there is at least one piece of evidence that does not belong to $\chi_i$, $\exists j . e_j \notin \chi_i$, should be equal no matter whether we base that belief on the original piece of metalevel evidence, before $e_q$ is taken out from $\chi_i$, or on a combination of the other two pieces of metalevel evidence $m_{\Delta\chi_i}(e_q \notin \chi_i)$ and $m_{\chi_i - \{e_q\}}(\exists j \neq q . e_j \notin (\chi_i - \{e_q\}))$, after $e_q$ is taken out from $\chi_i$, i.e.,

$$\text{Bel}_{\chi_i}(\exists j . e_j \notin \chi_i) = \text{Bel}_{\Delta\chi_i \oplus (\chi_i - \{e_q\})}(\exists j . e_j \notin \chi_i).$$

We may rewrite the original proposition

$$\exists j . e_j \notin \chi_i$$

as

$$(\exists j \neq q . e_j \notin \chi_i) \vee (e_q \notin \chi_i)$$

and as

$$(\exists j \neq q . e_j \notin (\chi_i - \{e_q\})) \vee (e_q \notin \chi_i).$$

Then, we have on the one hand, before $e_q$ is taken out from $\chi_i$,

$$\text{Bel}_{\chi_i}(\exists j . e_j \notin \chi_i) = m_{\chi_i}(\exists j . e_j \notin \chi_i) = c_i,$$

and on the other hand, if the metalevel evidence that there is at least one other piece of evidence $e_j$, $j \neq q$, that does not belong to $\chi_i - \{e_q\}$ is fused with the metalevel evidence that $e_q$ does not belong to $\chi_i$, Figure 2, we may also calculate the belief in $\exists j.e_j \notin \chi_i$ as

$$
\begin{aligned}
\mathrm{Bel}_{\Delta\chi_i \oplus (\chi_i - \{e_q\})}(\exists j.e_j \notin \chi_i) &= \mathrm{Bel}_{\Delta\chi_i \oplus (\chi_i - \{e_q\})}((\exists j \neq q.e_j \notin (\chi_i - \{e_q\})) \vee (e_q \notin \chi_i)) \\
&= \sum_{X \subseteq ((\exists j \neq q.e_j \notin (\chi_i - \{e_q\})) \vee (e_q \notin \chi_i))} m_{\Delta\chi_i \oplus (\chi_i - \{e_q\})}(X) \\
&= m_{\Delta\chi_i \oplus (\chi_i - \{e_q\})}((\exists j \neq q.e_j \notin (\chi_i - \{e_q\})) \wedge (e_q \notin \chi_i)) \\
&\quad + m_{\Delta\chi_i \oplus (\chi_i - \{e_q\})}(\exists j \neq q.e_j \notin (\chi_i - \{e_q\})) \\
&\quad + m_{\Delta\chi_i \oplus (\chi_i - \{e_q\})}(e_q \notin \chi_i) \\
&= m_{\chi_i - \{e_q\}}(\exists j \neq q.e_j \notin (\chi_i - \{e_q\})) \cdot m_{\Delta\chi_i}(e_q \notin \chi_i) \\
&\quad + m_{\chi_i - \{e_q\}}(\exists j \neq q.e_j \notin (\chi_i - \{e_q\})) \cdot [1 - m_{\Delta\chi_i}(e_q \notin \chi_i)] \\
&\quad + m_{\Delta\chi_i}(e_q \notin \chi_i) \cdot [1 - m_{\chi_i - \{e_q\}}(\exists j \neq q.e_j \notin (\chi_i - \{e_q\}))] \\
&= m_{\chi_i - \{e_q\}}(\exists j \neq q.e_j \notin (\chi_i - \{e_q\})) \\
&\quad + m_{\Delta\chi_i}(e_q \notin \chi_i) \cdot [1 - m_{\chi_i - \{e_q\}}(\exists j \neq q.e_j \notin (\chi_i - \{e_q\}))] \\
&= c_i^* + m_{\Delta\chi_i}(e_q \notin \chi_i) \cdot [1 - c_i^*].
\end{aligned}
$$

|  | $\exists j \neq q.e_j \notin (\chi_i - \{e_q\})$ | $\Theta$ |
|---|---|---|
| $e_q \notin \chi_i$ | $(\exists j \neq q.e_j \notin (\chi_i - \{e_q\})) \wedge (e_q \notin \chi_i)$ | $e_q \notin \chi_i$ |
| $\Theta$ | $\exists j \neq q.e_j \notin (\chi_i - \{e_q\})$ | $\Theta$ |

**Figure 2.** Combining $\mathrm{Bel}_{\chi_i - \{e_q\}}$ and $\mathrm{Bel}_{\Delta\chi_i}$.

Thus, we have derived a piece of metalevel evidence with a proposition stating that $e_q$ does not belong to $\chi_i$ from the variations in cluster conflict when $e_q$ was taken out from $\chi_i$:

$$
m_{\Delta\chi_i}(e_q \notin \chi_i) = \frac{c_i - c_i^*}{1 - c_i^*},
$$

$$
m_{\Delta\chi_i}(\Theta) = 1 - \frac{c_i - c_i^*}{1 - c_i^*} = \frac{1 - c_i}{1 - c_i^*}.
$$

If $e_q$ then is brought into another subset $\chi_k$ its conflict will increase from $c_k$ to $c_k^*$ where

$$\forall k \neq i.m_{\chi_k}(\exists j \neq q.e_j \notin \chi_k) = c_k,$$

$$\forall k \neq i.m_{\chi_k}(\Theta) = 1 - c_k.$$

and

$$\forall k \neq i.m_{\chi_k + \{e_q\}}(\exists j.e_j \notin (\chi_k + \{e_q\})) = c_k^*,$$

$$\forall k \neq i.m_{\chi_k + \{e_q\}}(\Theta) = 1 - c_k^*.$$

Thus we will also have metalevel evidence regarding every other subset $\chi_k$.

The increase in conflict when $e_q$ is brought into $\chi_k$ is interpreted as if there exists some metalevel evidence indicating that $e_q$ does not belong to $\chi_k + \{e_q\}$, i.e.,

$$m_{\Delta\chi_k}(e_q \notin (\chi_k + \{e_q\})),$$

$$m_{\Delta\chi_k}(\Theta).$$

Similarly to the last case, we may this time rewrite the new proposition

$$\exists j.e_j \notin (\chi_k + \{e_q\})$$

as

$$(\exists j \neq q.e_j \notin (\chi_k + \{e_q\})) \vee (e_q \notin (\chi_k + \{e_q\}))$$

and as

$$(\exists j \neq q.e_j \notin \chi_k) \vee (e_q \notin (\chi_k + \{e_q\}))$$

Reasoning in the same way as above, we state that

$$\forall k \neq i.\mathrm{Bel}_{\chi_k + \{e_q\}}(\exists j.e_j \notin (\chi_k + \{e_q\})) = \mathrm{Bel}_{\Delta\chi_k \oplus \chi_k}(\exists j.e_j \notin (\chi_k + \{e_q\}))$$

and find that on the one hand, after $e_q$ is brought into $\chi_k$

$$\forall k \neq i.\mathrm{Bel}_{\chi_k + \{e_q\}}(\exists j.e_j \notin (\chi_k + \{e_q\})) = m_{\chi_k + \{e_q\}}(\exists j.e_j \notin (\chi_k + \{e_q\})) = c_k^*$$

and on the other hand,

$$\forall k \neq i.\mathrm{Bel}_{\Delta\chi_k \oplus \chi_k}(\exists j.e_j \notin (\chi_k + \{e_q\}))$$
$$= \mathrm{Bel}_{\Delta\chi_k \oplus \chi_k}((\exists j \neq q.e_j \notin \chi_k) \vee (e_q \notin (\chi_k + \{e_q\})))$$
$$= \sum_{X \subseteq ((\exists j \neq q.e_j \notin \chi_k) \vee (e_q \notin (\chi_k + \{e_q\})))} m_{\Delta\chi_k \oplus \chi_k}(X)$$
$$= m_{\Delta\chi_k \oplus \chi_k}((\exists j \neq q.e_j \notin \chi_k) \wedge (e_q \notin (\chi_k + \{e_q\}))) + m_{\Delta\chi_k \oplus \chi_k}(\exists j \neq q.e_j \notin \chi_k)$$

$$+ m_{\Delta\chi_k \oplus \chi_k}(e_q \notin (\chi_k + \{e_q\})) = m_{\chi_k}(\exists j \neq q.e_j \in \chi_k) \cdot m_{\Delta\chi_k}(e_q \notin (\chi_k + \{e_q\}))$$

$$+ m_{\chi_k}(\exists j \neq q.e_j \in \chi_k) \cdot [1 - m_{\Delta\chi_k}(e_q \notin (\chi_k + \{e_q\}))]$$

$$+ m_{\Delta\chi_k}(e_q \notin (\chi_k + \{e_q\})) \cdot [1 - m_{\chi_k}(\exists j \neq q.e_j \in \chi_k)]$$

$$= m_{\chi_k}(\exists j \neq q.e_j \in \chi_k) + m_{\Delta\chi_k}(e_q \notin (\chi_k + \{e_q\})) \cdot [1 - m_{\chi_k}(\exists j \neq q.e_j \in \chi_k)]$$

$$= c_k + m_{\Delta\chi_k}(e_q \notin (\chi_k + \{e_q\})) \cdot [1 - c_k].$$

Thus, we have then derived a piece of metalevel evidence regarding each subset $\chi_k + \{e_q\}$, $k \neq i$, with a proposition stating that $e_q$ does not belong to the subset;

$$\forall k \neq i.m_{\Delta\chi_k}(e_q \notin (\chi_k + \{e_q\})) = \frac{c_k^* - c_k}{1 - c_k},$$

$$\forall k \neq i.m_{\Delta\chi_k}(\Theta) = 1 - \frac{c_k^* - c_k}{1 - c_k} = \frac{1 - c_k^*}{1 - c_k}.$$

## B. Evidence From Domain Conflict Variations

Since a piece of evidence from domain conflict is evidence against the entire partitioning it is less specific than a piece of evidence from cluster conflict. We will interpret the domain conflict as evidence that there exists at least one piece of evidence that does not belong to any of the $n$ first subsets, $n = |\chi|$, or if this particular piece of evidence was placed in a subset by itself, as evidence that it belongs to some of the other $n - 1$ subsets. This would indicate that the number of subsets is incorrect.

We will now study any changes in the domain conflict when we take out a piece of evidence $e_q$ from subset $\chi_i$.

When $|\chi_i| > 1$ we may not only put a piece of evidence $e_q$ that we have taken out from $\chi_i$ into another already existing subset, we may also put $e_q$ into a new subset $\chi_{n+1}$ by itself, assuming there are $n$ subsets, i.e., $\chi = \{\chi_1, ..., \chi_n\}$. This will change the domain conflict, $c_0$. Since the current partition minimizes the metaconflict function, we know that when the number of subsets increase we will get an increase in total conflict and Theorem 1 says that we will get a decrease in the nondomain part of the metaconflict function. Thus, we know that we must get an increase in the domain conflict. This increase in domain conflict is an indication that $e_q$ does not belong to an additional subset $\chi_{n+1}$.

Another way to receive a piece of evidence from the domain conflict is if $e_q$ is moved out from $\chi_i$ when $|\chi_i| = 1$. If $e_q$ is in a subset $\chi_i$ by itself and moved from $\chi_i$ to another already existing subset we may get either an increase or decrease in domain conflict. This is because both the total conflict and the nondomain part of the metaconflict function increases. Thus, we have two different situations in this case. If the domain conflict decreases when we move

$e_q$ out from $\chi_i$ this is interpreted as evidence that $e_q$ does not belongs to $\chi_i$, but if we receive an increase in domain conflict we will interpret this as evidence that $e_q$ does belong to $\chi_i$.

We choose to adopt a metarepresentation consisting of three individual representations for the domain conflict. The first representation interprets the domain conflict as evidence that there is at least one piece of evidence that does not belong to any of the subsets,

$$\exists j \forall k. e_j \notin \chi_k.$$

The second representation interprets the domain conflict as evidence that there is at least one subset to which no pieces of evidence belongs,

$$\exists k \forall j. e_j \notin \chi_k,$$

and the third as evidence that there is either at least one piece of evidence that does not belong to any of the subsets or there is at least one subset to which no pieces of evidence belongs, but not both at the same time,

$$[(\exists j \forall k. e_j \notin \chi_k) \wedge (\neg \exists k \forall j. e_j \notin \chi_k)] \vee [(\neg \exists j \forall k. e_j \notin \chi_k) \wedge (\exists k \forall j. e_j \notin \chi_k)].$$

Each representation has its own characteristics. The first and the second are consistent with a situation where the domain conflict increases when the number of subsets increase and decreases when the number of subsets decrease. The third representation behaves in the opposite way. The three representations above correspond to these three different situations when the number of subsets is changed.

The first representation corresponds to the situation when one piece of evidence $e_q$ belongs to a subset $\chi_i$, $|\chi_i| > 1$, and it is moved from $\chi_i$ to $\chi_{n+1}$. This increases both the domain conflict and the number of subsets. The second representation is not in accordance with this situation and the third is not even consistent with the situation. The second representation corresponds to the situation when one piece of evidence $e_q$ belongs to a subset $\chi_i$, $|\chi_i| = 1$, and it is moved from $\chi_i$ to one of the other $n - 1$ subsets while we receive a decrease in domain conflict. Here, the first representation is not in accordance with this situation and the third is not consistent. The third representation corresponds to the last situation when one piece of evidence $e_q$ belongs to a subset $\chi_i$, $|\chi_i| = 1$, and it is moved from $\chi_i$ to one of the other $n - 1$ subsets while we receive an increase in domain conflict. In this situation the first and second representations are not consistent.

Thus, the actual representation to be used can be chosen from the metarepresentation by the current situation. Let us now see what can be derived about our piece of evidence of interest, $e_q$.

### 1.   When $e_q \in \chi_i$ and $|\chi_i| > 1$

Let us analyze the case where we move $e_q$ from $\chi_i$ to $\chi_{n+1}$. The domain conflict before $e_q$ is moved to $\chi_{n+1}$ is interpreted as

$$\exists j \forall k \neq n + 1. e_j \notin \chi_k.$$

When $e_q \in \chi_i$ and $|\chi_i| > 1$ we may move out $e_q$ from $\chi_i$ without changing the domain conflict, but we will get an increase in the domain conflict if we move $e_q$ to a subset by itself; $\chi_{n+1}$.

If $e_q$ is taken out from $\chi_i$, without being moved to $\chi_{n+1}$, and is temporarily disregarded from the analysis we will still have $n$ subsets in this new situation since $|\chi_i|$ was greater than one. The domain conflict, which is unchanged by this removal and equal to $c_0$, may be interpreted in this case as

$$\exists j \neq q \forall k \neq n+1 . e_j \in \chi_k.$$

Thus, when $e_q$ is taken out from $\chi_i$, we can refine the bpa regarding the domain conflict to

$$m_\chi(\exists j \neq q \forall k \neq n+1 . e_j \notin \chi_k) = c_0,$$
$$m_\chi(\Theta) = 1 - c_0.$$

If $e_q$ then is moved to $\chi_{n+1}$ the increase in domain conflict is evidence that $e_q$ does not belong to $\chi_{n+1}$,

$$m_{\Delta\chi}(e_q \notin \chi_{n+1}),$$
$$m_{\Delta\chi}(\Theta),$$

and the new domain conflict that we receive indicate that there must now be at least one piece of evidence that does not belong to any of the $n + 1$ subsets,

$$m_{\chi + \{\chi_{n+1}\}}(\exists j \forall k . e_j \notin \chi_k) = c_0^*,$$
$$m_{\chi + \{\chi_{n+1}\}}(\Theta) = 1 - c_0^*.$$

We will derive $m_{\Delta\chi}(e_q \notin \chi_{n+1})$ by stating that

$$\mathrm{Bel}_{\chi + \{\chi_{n+1}\}}(\exists j \forall k . e_j \notin \chi_k) = \mathrm{Bel}_{\Delta\chi \oplus \chi}(\exists j \forall k . e_j \notin \chi_k).$$

After $e_q$ is moved to $\chi_{n+1}$, we may rewrite the proposition

$$\exists j \forall k . e_j \notin \chi_k$$

as

$$[\exists j \neq q((\forall k \neq n+1 . e_j \notin \chi_k) \wedge (e_j \notin \chi_{n+1}))]$$
$$\vee [(\forall k \neq n+1 . e_q \notin \chi_k) \wedge (e_q \notin \chi_{n+1})]$$

and as

$$(\exists j \neq q \forall k \neq n+1 . e_j \notin \chi_k) \vee (e_q \notin \chi_{n+1})$$

since we know it is true that $e_j \notin \chi_{n+1}$ for some other pieces of evidence than $e_q$ and it is true that $e_q \notin \chi_k$ for all other subsets than $\chi_{n+1}$, since $e_q$ is now in $\chi_{n+1}$.

Then, on the one hand we have

$$\mathrm{Bel}_{\chi + \{\chi_{n+1}\}}(\exists j \forall k. e_j \notin \chi_k) = m_{\chi + \{\chi_{n+1}\}}(\exists j \forall k. e_j \notin \chi_k) = c_0^*$$

and on the other hand we can calculate

$$
\begin{aligned}
\mathrm{Bel}_{\Delta\chi \oplus \chi}(\exists j \forall k. e_j \notin \chi_k) &= \mathrm{Bel}_{\Delta\chi \oplus \chi}((\exists j \neq q \forall k \neq n+1. e_j \notin \chi_k) \vee (e_q \notin \chi_{n+1})) \\
&= \sum_{X \subseteq ((\exists j \neq q \forall k \neq n+1. e_j \notin \chi_k) \vee (e_q \notin \chi_{n+1}))} m_{\Delta\chi \oplus \chi}(X) \\
&= m_{\Delta\chi \oplus \chi}((\exists j \neq q \forall k \neq n+1. e_j \notin \chi_k) \wedge (e_q \notin \chi_{n+1})) \\
&\quad + m_{\Delta\chi \oplus \chi}(\exists j \neq q \forall k \neq n+1. e_j \notin \chi_k) + m_{\Delta\chi \oplus \chi}(e_q \notin \chi_{n+1}) \\
&= m_{\chi}(\exists j \neq q \forall k \neq n+1. e_j \notin \chi_k) \cdot m_{\Delta\chi}(e_q \notin \chi_{n+1}) \\
&\quad + m_{\chi}(\exists j \neq q \forall k \neq n+1. e_j \notin \chi_k) \cdot [1 - m_{\Delta\chi}(e_q \notin \chi_{n+1})] \\
&\quad + m_{\Delta\chi}(e_q \notin \chi_{n+1}) \cdot [1 - m_{\chi}(\exists j \neq q \forall k \neq n+1. e_j \notin \chi_k)] \\
&= m_{\chi}(\exists j \neq q \forall k \neq n+1. e_j \notin \chi_k) + m_{\Delta\chi}(e_q \notin \chi_{n+1}) \\
&\quad \cdot [1 - m_{\chi}(\exists j \neq q \forall k \neq n+1. e_j \notin \chi_k)] \\
&= c_0 + m_{\Delta\chi}(e_q \notin \chi_{n+1}) \cdot [1 - c_0].
\end{aligned}
$$

Thus, we get

$$m_{\Delta\chi}(e_q \notin \chi_{n+1}) = \frac{c_0^* - c_0}{1 - c_0}$$

a piece of evidence indicating that $e_q$ does not belong to $\chi_{n+1}$.

## 2.   When $e_q \in \chi_i$, $|\chi_i| = 1$ and $c_0 > c_0^*$

Let us now study the situation when $e_q$ is in a subset by itself. Before $e_q$ is moved out from $\chi_i$,

$$\exists k \forall j. e_j \notin \chi_k$$

represents the domain conflict with the $n$ subsets. Thus, we may refine the bpa regarding domain conflict in this situation to

$$
\begin{aligned}
m_{\chi}(\exists k \forall j. e_j \notin \chi_k) &= c_0, \\
m_{\chi}(\Theta) &= 1 - c_0.
\end{aligned}
$$

If we take out $e_q$ from $\chi_i$, without moving it to any already existing subset, and temporarily disregard it from the analysis, we have $n - 1$ remaining subsets and get a decrease in the domain conflict. This decrease in domain conflict is interpreted as evidence that $e_q$ does not belong to $\chi_i$,

$$m_{\Delta\chi}(e_q \notin \chi_i),$$

$$m_{\Delta\chi}(\Theta).$$

The remaining domain conflict indicate that there is now at least one other subset $\chi_k$, $k \neq i$, that does not contain any pieces of evidence $e_j$, $j \neq q$,

$$m_{\chi - \{\chi_i\}}(\exists k \neq i \forall j \neq q.e_j \notin \chi_k) = c_0^*,$$

$$m_{\chi - \{\chi_i\}}(\Theta) = 1 - c_0^*.$$

As before, we will derive $m_{\Delta\chi}(e_q \notin \chi_i)$ by stating that

$$\text{Bel}_\chi(\exists k \forall j.e_j \notin \chi_k) = \text{Bel}_{\Delta\chi \oplus (\chi - \{\chi_i\})}(\exists k \forall j.e_j \notin \chi_k).$$

Before $e_q$ is taken out from $\chi_i$ we may rewrite the proposition

$$\exists k \forall j.e_j \notin \chi_k$$

as

$$[\exists k \neq i((\forall j \neq q.e_j \notin \chi_k) \wedge (e_q \notin \chi_k))] \vee [(\forall j \neq q.e_j \notin \chi_i) \wedge (e_q \notin \chi_i)]$$

and as

$$(\exists k \neq i \forall j \neq q.e_j \notin \chi_k) \vee (e_q \notin \chi_i).$$

since $e_q \notin \chi_k$ for some $k \neq i$ and $e_j \notin \chi_i$ for all other pieces of evidence than $e_q$, since $e_q$ is still in $\chi_i$.

Similarly to the previous case we have

$$\text{Bel}_\chi(\exists k \forall j.e_j \notin \chi_k) = m_\chi(\exists k \forall j.e_j \notin \chi_k) = c_0$$

and may calculate

$$\text{Bel}_{\Delta\chi \oplus (\chi - \{\chi_i\})}(\exists k \forall j.e_j \notin \chi_k) = \text{Bel}_{\Delta\chi \oplus (\chi - \{\chi_i\})}((\exists k \neq i \forall j \neq q.e_j \notin \chi_k) \vee (e_q \notin \chi_i))$$

$$= \sum_{X \subseteq ((\exists k \neq i \forall j \neq q.e_j \notin \chi_k) \vee (e_q \notin \chi_i))} m_{\Delta\chi \oplus (\chi - \{\chi_i\})}(X)$$

$$= m_{\Delta\chi \oplus (\chi - \{\chi_i\})}((\exists k \neq i \forall j \neq q.e_j \notin \chi_k) \wedge (e_q \notin \chi_i))$$

$$+ m_{\Delta\chi \oplus (\chi - \{\chi_i\})}(\exists k \neq i \forall j \neq q.e_j \notin \chi_k) + m_{\Delta\chi \oplus (\chi - \{\chi_i\})}(e_q \notin \chi_i)$$

$$= m_{\chi - \{\chi_i\}}(\exists k \neq i \forall j \neq q.e_j \notin \chi_k) \cdot m_{\Delta\chi}(e_q \notin \chi_i)$$

$$+ m_{\chi - \{\chi_i\}}(\exists k \neq i \forall j \neq q.e_j \notin \chi_k) \cdot [1 - m_{\Delta\chi}(e_q \notin \chi_i)]$$

$$+ m_{\Delta\chi}(e_q \notin \chi_i) \cdot [1 - m_{\chi - \{\chi_i\}}(\exists k \neq i \forall j \neq q.e_j \notin \chi_k)]$$

$$= m_{\chi - \{\chi_i\}}(\exists k \neq i \forall j \neq q.e_j \notin \chi_k) + m_{\Delta\chi}(e_q \notin \chi_i)$$

$$\cdot [1 - m_{\chi - \{\chi_i\}}(\exists k \neq i \forall j \neq q.e_j \notin \chi_k)]$$

$$= c_0^* + m_{\Delta\chi}(e_q \notin \chi_i) \cdot [1 - c_0^*].$$

Thus, when $c_0 > c_0^*$ we have

$$m_{\Delta\chi}(e_q \notin \chi_i) = \frac{c_0 - c_0^*}{1 - c_0^*}.$$

*3. When $e_q \in \chi_i$, $|\chi_i| = 1$ and $c_0 < c_0^*$*

A completely different situation occurs when $c_0 < c_0^*$. This is the case when we choose to represent the domain conflict as an exclusive-OR of two propositions,

$$[(\exists j \forall k.e_j \notin \chi_k) \wedge (\neg\exists k \forall j.e_j \notin \chi_k)] \vee [(\neg\exists j \forall k.e_j \notin \chi_k) \wedge (\exists k \forall j.e_j \notin \chi_k)],$$

one stating that there is at least one piece of evidence that does not belong to any of the $n$ subsets and the other one stating that there is at least one subset that does not contain any pieces of evidence. Thus, we can refine the bpa regarding domain conflict to

$$m_\chi\!\left(\begin{array}{l}[(\exists j \forall k.e_j \notin \chi_k) \wedge (\neg\exists k \forall j.e_j \notin \chi_k)] \\ \vee [(\neg\exists j \forall k.e_j \notin \chi_k) \wedge (\exists k \forall j.e_j \notin \chi_k)]\end{array}\right) = c_0,$$

$$m_\chi(\Theta) = 1 - c_0.$$

If we take out $e_q$ from $\chi_i$ as in the previous case, without moving it to any already existing subset, and temporarily disregard it from the analysis, we have $n - 1$ remaining subsets since $|\chi_i|$ was equal to one and get an increase in the domain conflict. This increase in domain conflict will be interpreted as evidence that $e_q$ belongs to $\chi_i$,

$$m_{\Delta\chi}(e_q \in \chi_i),$$

$$m_{\Delta\chi}(\Theta).$$

The remaining domain conflict indicate that there is now at least one other subset $\chi_k$, $k \neq i$, that does not contain any pieces of evidence $e_j$, $j \neq q$, or that there is at least one piece of evidence $e_j$, $j \neq q$, that does not belong to any of the $n - 1$ subsets $\chi_k$, $k \neq i$,

$$m_{\chi - \{\chi_i\}}\!\left(\begin{array}{l}((\exists j \neq q \forall k \neq i.e_j \notin \chi_k) \wedge (\neg\exists k \neq i \forall j \neq q.e_j \in \chi_k)) \\ \vee ((\neg\exists j \neq q \forall k \neq i.e_j \in \chi_k) \wedge (\exists k \neq i \forall j \neq q.e_j \notin \chi_k))\end{array}\right) = c_0^*,$$

$$m_{\chi - \{\chi_i\}}(\Theta) = 1 - c_0^*.$$

We will derive $m_{\Delta\chi}(\cdot)$ by stating that

$$\mathrm{Bel}_\chi\!\left(\begin{array}{l}[(\exists j \forall k.e_j \notin \chi_k) \wedge (\neg\exists k \forall j.e_j \notin \chi_k)] \\ \vee [(\neg\exists j \forall k.e_j \notin \chi_k) \wedge (\exists k \forall j.e_j \notin \chi_k)]\end{array}\right)$$

$$= \mathrm{Bel}_{\Delta\chi \oplus (\chi - \{\chi_i\})}\!\left(\begin{array}{l}[(\exists j \forall k.e_j \notin \chi_k) \wedge (\neg\exists k \forall j.e_j \notin \chi_k)] \\ \vee [(\neg\exists j \forall k.e_j \notin \chi_k) \wedge (\exists k \forall j.e_j \notin \chi_k)]\end{array}\right).$$

The domain conflict, represented by

$$[(\exists j \forall k.e_j \notin \chi_k) \wedge (\neg \exists k \forall j.e_j \notin \chi_k)] \vee [(\neg \exists j \forall k.e_j \notin \chi_k) \wedge (\exists k \forall j.e_j \notin \chi_k)],$$

can be rewritten as

$$[((\exists j \neq q \forall k \neq i.e_j \notin \chi_k) \vee (e_q \notin \chi_i)) \wedge ((\forall k \neq i \exists j \neq q.e_j \in \chi_k) \wedge (e_q \in \chi_i))]$$
$$\vee [((\forall j \neq q \exists k \neq i.e_j \in \chi_k) \wedge (e_q \in \chi_i)) \wedge ((\exists k \neq i \forall j \neq q.e_j \notin \chi_k) \vee (e_q \notin \chi_i))],$$

by using the simplifications of the previous two sections and further simplified as

$$((\exists j \neq q \forall k \neq i.e_j \notin \chi_k) \wedge (\forall k \neq i \exists j \neq q.e_j \in \chi_k) \wedge (e_q \in \chi_i))$$
$$\vee ((\forall j \neq q \exists k \neq i.e_j \in \chi_k) \wedge (\exists k \neq i \forall j \neq q.e_j \notin \chi_k) \vee (e_q \in \chi_i)),$$

and finally restated as

$$[((\exists j \neq q \forall k \neq i.e_j \notin \chi_k) \wedge (\neg \exists k \neq i \forall j \neq q.e_j \in \chi_k))$$
$$\vee ((\neg \exists j \neq q \forall k \neq i.e_j \in \chi_k) \wedge (\exists k \neq i \forall j \neq q.e_j \in \chi_k))] \wedge (e_q \in \chi_i)$$

where the first part is interpreted as the domain conflict before $e_q$ is moved from $\chi_i$ and the second a proposition stating that $e_q$ belongs to $\chi_i$.

Then, on the one hand we have

$$\text{Bel}_\chi \begin{pmatrix} [(\exists j \forall k.e_j \notin \chi_k) \wedge (\neg \exists k \forall j.e_j \notin \chi_k)] \\ \vee [(\neg \exists j \forall k.e_j \notin \chi_k) \wedge (\exists k \forall j.e_j \notin \chi_k)] \end{pmatrix}$$
$$= m_\chi \begin{pmatrix} [(\exists j \forall k.e_j \notin \chi_k) \wedge (\neg \exists k \forall j.e_j \notin \chi_k)] \\ \vee [(\neg \exists j \forall k.e_j \notin \chi_k) \wedge (\exists k \forall j.e_j \notin \chi_k)] \end{pmatrix} = c_0$$

and on the other hand we can calculate

$$\text{Bel}_{\Delta\chi \oplus (\chi - \{\chi_i\})} \begin{pmatrix} [(\exists j \forall k.e_j \notin \chi_k) \wedge (\neg \exists k \forall j.e_j \notin \chi_k)] \\ \vee [(\neg \exists j \forall k.e_j \notin \chi_k) \wedge (\exists k \forall j.e_j \notin \chi_k)] \end{pmatrix}$$

$$= \text{Bel}_{\Delta\chi \oplus (\chi - \{\chi_i\})} \begin{pmatrix} [((\exists j \neq q \forall k \neq i.e_j \notin \chi_k) \wedge (\neg \exists k \neq i \forall j \neq q.e_j \in \chi_k)) \\ \vee ((\neg \exists j \neq q \forall k \neq i.e_j \in \chi_k) \wedge (\exists k \neq i \forall j \neq q.e_j \notin \chi_k))] \\ \wedge (e_q \in \chi_i) \end{pmatrix}$$

$$= \sum_{X \subseteq \begin{pmatrix} [((\exists j \neq q \forall k \neq i.e_j \in \chi_k) \wedge (\neg \exists k \neq i \forall j \neq q.e_j \in \chi_k)) \\ \vee ((\neg \exists j \neq q \forall k \neq i.e_j \in \chi_k) \wedge (\exists k \neq i \forall j \neq q.e_j \notin \chi_k))] \wedge (e_q \in \chi_i) \end{pmatrix}} m_{\Delta\chi \oplus (\chi - \{\chi_i\})}(X)$$

$$= m_{\Delta\chi \oplus (\chi - \{\chi_i\})} \begin{pmatrix} [((\exists j \neq q \forall k \neq i.e_j \notin \chi_k) \wedge (\neg \exists k \neq i \forall j \neq q.e_j \in \chi_k)) \\ \vee ((\neg \exists j \neq q \forall k \neq i.e_j \in \chi_k) \wedge (\exists k \neq i \forall j \neq q.e_j \notin \chi_k))] \\ \wedge (e_q \in \chi_i) \end{pmatrix}$$

$$= m_{\chi - \{\chi_i\}} \left( \begin{array}{l} [((\exists j \neq q \forall k \neq i.e_j \notin \chi_k) \wedge (\neg \exists k \neq i \forall j \neq q.e_j \in \chi_k)) \\ \vee ((\neg \exists j \neq q \forall k \neq i.e_j \in \chi_k) \wedge (\exists k \neq i \forall j \neq q.e_j \notin \chi_k))] \end{array} \right)$$

$$\cdot m_{\Delta\chi}(e_q \in \chi_i) = c_0^* \cdot m_{\Delta\chi}(e_q \in \chi_i).$$

Thus, if $c_0 < c_0^*$ then

$$m_{\Delta\chi}(e_q \in \chi_i) = \frac{c_0}{c_0^*}$$

is derived as our piece of evidence.

## C.   Summary of Evidence

In summary we have the following pieces of evidence, (for all $i$ when $e_q \in \chi_i$ we have $m(e_q \notin \chi_j) = \dots$),

$$\forall i, e_q \in \chi_i . m(e_q \notin \chi_j) = \begin{cases} \dfrac{c_0^* - c_0}{1 - c_0}, j = n+1, |\chi_i| > 1 \\[2ex] \dfrac{c_i - c_i^*}{1 - c_i^*}, j = i, |\chi_i| > 1 \\[2ex] \dfrac{c_0 - c_0^*}{1 - c_0^*}, j = i, |\chi_i| = 1, c_0 > c_0^* \\[2ex] \dfrac{c_j^* - c_j}{1 - c_j}, \text{otherwise} \end{cases}$$

and

$$\forall i, e_q \in \chi_i . m(e_q \in \chi_i) = \frac{c_0}{c_0^*}, |\chi_i| = 1, c_0 < c_0^*.$$

## APPENDIX II:   SPECIFYING EVIDENCE

We may now specify any original piece of evidence by combining all evidence from conflict variations regarding this particular piece of evidence. Then we may calculate the belief and plausibility for each subset that this particular piece of evidence belongs to the subset. The belief that it belongs to a subset will be zero, except when $e_q \in \chi_i$, $|\chi_i| = 1$ and $c_0 < c_0^*$, since every proposition regarding this piece of evidence then states that it does not belong to some subset.

In combining all pieces of evidence regarding an original piece of evidence we may receive support for a proposition stating that this piece of evidence does not belong to any of the subsets and cannot be put into a subset by itself. Since this is impossible, the statement is false and its support is the conflict in Dempster's rule. The statement that a piece of evidence does not belong any-where implies that it is false. Thus, we may interpret the conflict as support for this piece of evidence being false.

## A. Combining Evidence About Evidence

### 1. When $e_q \in \chi_i$ and $|\chi_i| > 1$

Let us assume that a piece of evidence, $e_q$, is in $\chi_i$ and $|\chi_i| > 1$. When we combine all pieces of evidence regarding $e_q$ this results in a new basic probability assignment with

$$\forall \chi^* . m^*(e_q \notin (\vee \chi^*)) = \prod_{\chi_j \in \chi^*} m(e_q \in \chi_j) \cdot \prod_{\chi_j \in (\chi - \chi^*)} [1 - m(e_q \notin \chi_j)]$$

where $\chi^* \in 2^\chi$, $\chi = \{\chi_1, ..., \chi_{n+1}\}$ and $\vee \chi^*$ is the disjunction of all elements in $\chi^*$.

From the new bpa we can calculate the conflict. The only statement that is false is the statement that $e_q \notin (\vee \chi)$, i.e., that $\forall j . e_q \notin \chi_j$.

Thus, the conflict becomes

$$k = m^*(e_q \notin (\vee \chi)) = \prod_{j=1}^{n+1} m(e_q \notin \chi_j) = \frac{c_0^* - c_0}{1 - c_0} \cdot \prod_{j=1}^{n} \frac{c_j^* - c_j}{1 - c_j}.$$

When calculating belief and plausibility that $e_q$ belongs to some subset other than $\chi_{n+1}$ we have

$$\forall k \neq n+1 . \mathrm{Bel}(e_q \in \chi_k) = 1 - \sum_{X \subseteq (e_q \in \chi_k)} m(X) = 0$$

and

$$\forall k \neq n+1 . \mathrm{Pls}(e_q \in \chi_k) = 1 - \mathrm{Bel}(e_q \notin \chi_k) = 1 - \sum_{X \subseteq (e_q \notin \chi_k)} m(X)$$

$$= 1 - \frac{1}{1-k} \cdot m(e_q \notin \chi_k) \cdot \left[ 1 - \prod_{\substack{j=1 \\ \neq k}}^{n+1} m(e_q \notin \chi_j) \right]$$

$$= 1 - \frac{m(e_q \notin \chi_k) - \prod_{j=1}^{n+1} m(e_q \notin \chi_j)}{1 - \prod_{j=1}^{n+1} m(e_q \notin \chi_j)} = \frac{1 - m(e_q \notin \chi_k)}{1 - \prod_{j=1}^{n+1} m(e_q \notin \chi_j)}$$

$$= \frac{1 - \frac{c_k^* - c_k}{1 - c_k}}{1 - \frac{c_0^* - c_0}{1 - c_0} \cdot \prod_{j=1}^{n} \frac{c_j^* - c_j}{1 - c_j}}$$

while for the subset $\chi_{n+1}$ we have

$$\mathrm{Bel}(e_q \in \chi_{n+1}) = 1 - \sum_{X \subseteq (e_q \in \chi_{n+1})} m(X) = 0$$

and

$$\text{Pls}(e_q \in \chi_{n+1}) = 1 - \text{Bel}(e_q \notin \chi_{n+1}) = 1 - \sum_{X \subseteq (e_q \notin \chi_{n+1})} m(X)$$

$$= 1 - \frac{1}{1-k} \cdot m(e_q \notin \chi_{n+1}) \cdot \left[ 1 - \prod_{j=1}^{n} m(e_q \notin \chi_j) \right]$$

$$= 1 - \frac{m(e_q \notin \chi_{n+1}) - \prod_{j=1}^{n+1} m(e_q \notin \chi_j)}{1 - \prod_{j=1}^{n+1} m(e_q \notin \chi_j)} = \frac{1 - m(e_q \notin \chi_{n+1})}{1 - \prod_{j=1}^{n+1} m(e_q \notin \chi_j)}$$

$$= \frac{1 - \dfrac{c_0^* - c_0}{1 - c_0}}{1 - \dfrac{c_0^* - c_0}{1 - c_0} \cdot \prod_{j=1}^{n} \dfrac{c_j^* - c_j}{1 - c_j}}.$$

*2.    When $e_q \in \chi_i$, $|\chi_i| = 1$ and $c_0 > c_0^*$*

In this situation the domain conflict variation appeared in the *i*th piece of evidence instead of the $n + 1$th. When we combine the available evidence we get a new basic probability assignment with

$$\forall \chi^*.m^*(e_q \notin (\ \lor \chi^*)) = \prod_{\chi_j \in \chi^*} m(e_q \notin \chi_j) \cdot \prod_{\chi_j \in (\chi - \chi^*)} [1 - m(e_q \notin \chi_j)]$$

but here $\chi^* \in 2^\chi$ where $\chi = \{\chi_1, ..., \chi_n\}$ .

With no evidence from a $n + 1$th subset and domain conflict variation in the *i*th piece of evidence we have a slight change in the calculation of conflict,

$$k = m^*(e_q \notin (\ \lor \chi)) = \prod_{j=1}^{n} m(e_q \notin \chi_j) = \frac{c_0 - c_0^*}{1 - c_0} \cdot \prod_{\substack{j=1 \\ \neq i}}^{n} \frac{c_j^* - c_j}{1 - c_j}$$

and in the calculation of plausibility. For subsets except $\chi_i$ we get

$$\forall k \neq i.\text{Bel}(e_q \in \chi_k) = 0,$$

$$\forall k \neq i.\text{Pls}(e_q \in \chi_k) = 1 - \text{Bel}(e_q \notin \chi_k) = 1 - \sum_{X \subseteq (e_q \notin \chi_k)} m(X)$$

$$= \frac{1 - m(e_q \notin \chi_k)}{1 - \prod_{j=1}^{n} m(e_q \notin \chi_j)} = \frac{1 - \dfrac{c_k^* - c_k}{1 - c_k}}{1 - \dfrac{c_0 - c_0^*}{1 - c_0^*} \cdot \prod_{\substack{j=1 \\ \neq i}}^{n} \dfrac{c_j^* - c_j}{1 - c_j}}$$

and for $\chi_i$

$$\mathrm{Bel}(e_q \in \chi_i) = 0,$$

$$\mathrm{Pls}(e_q \in \chi_i) = 1 - \mathrm{Bel}(e_q \notin \chi_i) = 1 - \sum_{X \subseteq (e_q \notin \chi_i)} m(X)$$

$$= \frac{1 - m(e_q \notin \chi_i)}{1 - \prod\limits_{j=1}^{n} m(e_q \notin \chi_j)} = \frac{1 - \dfrac{c_0^* - c_0}{1 - c_0}}{1 - \dfrac{c_0^* - c_0}{1 - c_0} \cdot \prod\limits_{\substack{j=1 \\ \neq i}}^{n} \dfrac{c_j^* - c_j}{1 - c_j}}.$$

### 3. When $e_q \in \chi_i$, $|\chi_i| = 1$ and $c_0 < c_0^*$

The increase in domain conflict when $e_q$ was moved out from $\chi_i$ introduced a new type of evidence supporting that $e_q$ belongs to $\chi_i$. This changes the resulting bpa from the previous situations when we combine all pieces of evidence regarding $e_q$. Our new bpa is

$$\forall \chi^* . m^*(e_q \notin (\vee \chi^*))$$

$$= [1 - m(e_q \in \chi_i)] \cdot \prod_{\chi_j \in \chi^*} m(e_q \notin \chi_j) \cdot \prod_{\chi_j \in (\chi^{-i} - \chi^*)} [1 - m(e_q \notin \chi_j)]$$

and

$$m^*(e_q \in (\vee \chi^{-i})) = m(e_q \in \chi_i) + [1 - m(e_q \in \chi_i)] \cdot \prod_{\chi_j \in \chi^{-i}} m(e_q \notin \chi_j)$$

where $\chi^* \in 2^{\chi^{-i}} - \{\chi^{-i}\}$, $\chi^{-i} = \chi - \{\chi_i\}$ and $\chi = \{\chi_1, ..., \chi_n\}$.

Since we did not have a piece of evidence indicating that $e_q$ did not belong to $\chi_i$ we will never have any support for the impossible statement that $e_q$ does not belong anywhere in the new bpa. Thus, we will always get a zero conflict when combining these pieces of evidence;

$$k = m^*(e_q \notin (\vee \chi)) = 0.$$

When calculating belief and plausibility for any subset other than $\chi_i$ we get

$$\forall k \neq i . \mathrm{Bel}(e_q \in \chi_k) = 0$$

and

$$\forall k \neq i . \mathrm{Pls}(e_q \in \chi_k) = 1 - \mathrm{Bel}(e_q \notin \chi_k) = 1 - \sum_{X \subseteq (e_q \notin \chi_k)} m(X)$$

$$= 1 - [m(e_q \in \chi_i) + \{1 - m(e_q \in \chi_i)\} \cdot m(e_q \notin \chi_k)]$$

$$= [1 - m(e_q \in \chi_i)] \cdot [1 - m(e_q \notin \chi_k)]$$

$$= \left(1 - \frac{c_0}{c_0^*}\right) \cdot \left(1 - \frac{c_k^* - c_k}{1 - c_k}\right)$$

and for $\chi_i$:

$$\text{Bel}(e_q \in \chi_i) = m(e_q \in \chi_i) + [1 - m(e_q \in \chi_i)] \cdot \prod_{\chi_j \in \chi^{-i}} m(e_q \notin \chi_j)$$

$$= \frac{c_0}{\overset{*}{c_0}} + \left(1 - \frac{c_0}{\overset{*}{c_0}}\right) \cdot \prod_{\substack{j=1 \\ \neq i}}^{n} \frac{c_j^* - c_j}{1 - c_j}.$$

and

$$\text{Pls}(e_q \in \chi_i) = 1 - \text{Bel}(e_q \notin \chi_i) = 1 - \sum_{X \subseteq (e_q \in \chi_i)} m(X) = 1$$

because of the lack of evidence against that $e_q$ belongs to the $i$th subset.

## B.    The Evidence Specified

With plausibilities for all propositions that the evidence is referring to some particular subset we may now make a partial specification of each piece of evidence. That is, we will have an "evidential interval" of belief and plausibility for each possible subset. Since $e_q$ belong to $\chi_i$ as a result of the iterative partitioning of all pieces of evidence, there was the least support against this and thus we will have the highest plausibility in favor of the proposition that $e_q$ is referring to subset $i$. A piece of evidence nonspecific with regard to which event it is referring to may then be specified from

> evidence $q$:
>     proposition:
>         action part: $A_f, A_g, ..., A_h$
>         event part: $E_i, E_j, ..., E_k$
> $m(A_f) = p_f$
> $m(A_g) = p_g$
> ...
> $m(A_h) = p_h$
> $m(\Theta) = 1 - p_f - p_g - ... - p_h$

to

> evidence $q$:
>     proposition:
>         action part: $A_f, A_g, ..., A_h$
>         event part: $[\text{Bel}(e_q \in \chi_i), \text{Pls}(e_q \in \chi_i)] / E_i, \ [0, \text{Pls}(e_q \in \chi_j)] / E_j, ...,$
>                   $[0, \text{Pls}(e_q \in \chi_k)] / E_k$
> $m(A_f) = p_f$
> $m(A_g) = p_g$
> ...
> $m(A_h) = p_h$
> $m(\Theta) = 1 - p_f - p_g - ... - p_h$